\documentclass[review]{elsarticle}

\usepackage{lineno,hyperref}
\modulolinenumbers[5]

\journal{Physica D}

\usepackage{numcompress}

\usepackage{graphicx}
\usepackage{amsmath}
\usepackage{gensymb}
\usepackage{enumitem}
\usepackage{algcompatible}

\begin{document}
\begin{frontmatter}

\title{Adding Filters to Improve Reservoir Computer Performance}
\author{T. L. Carroll}
\address{Code 6392, US Naval Research Lab}
\address{Washington, DC 20375 USA}
\ead{thomas.carroll@nrl.navy.mil}

\date{\today}

\begin{abstract}
Reservoir computers are a type of neuromorphic computer that may be built with analog hardware, potentially creating powerful computers that are small, light and consume little power. Typically a reservoir computer is build by connecting together a set of nonlinear nodes into a network; connecting the nonlinear nodes may be difficult or expensive, however. This work shows how a reservoir computer may be expanded by adding functions to its output. The particular functions described here are linear filters, but other functions are possible. The design and construction of linear filters is well known, and such filters may be easily implemented in hardware such as field programmable gate arrays (FPGA's). The effect of adding filters on the reservoir computer performance is simulated for a signal fitting problem, a prediction problem and a signal classification problem.
\end{abstract}


\end{frontmatter}

\maketitle

\section{Introduction}
A reservoir computer is a high dimensional dynamical system that may be used to do computation \cite{jaeger2001, natschlaeger2002}. The reservoir computer by itself will evolve to a stable fixed point; in use the reservoir computer is driven by an input signal $s(n)$. The reservoir computer is synchronized in the general sense to the input signal. meaning that the reservoir computer will follow the same trajectory every time it is driven with the same input signal (after an initial transient). To train a reservoir computer, a number of time series signals are extracted and used to do a linear fit to a training signal. The fit coefficients are the output of the training process. To use the reservoir computer for a computation, the same dynamical system is driven with a different input signal and a linear combination is made from the output signals using the coefficients found during the training process. An example of training and testing would be using the $x, y$ and $z$ signals from the Lorenz chaotic system as the input signal and creating a linear combination of reservoir signals to fit the Lorenz $z$ signal \cite{lu2017}. In the computation stage, the reservoir would be driven by the signals $x', y'$ and $z'$ from the Lorenz system with different initial conditions, and the linear combination made from the training coefficients and the reservoir signals would be a close fit to the corresponding $z'$ signal.

Typically a reservoir computer is build by linking together a set of nonlinear nodes in a directed network. A reservoir computer is similar to a recurrent neural network, but the connections between nodes do not change in a reservoir computer. As a result, reservoir computers may be built as analog systems.  Examples of analog reservoir computers so far include photonic systems \cite{larger2012, van_der_sande2017}, analog circuits \cite{schurmann2004}, mechanical systems \cite{dion2018} and  field programmable gate arrays \cite{canaday2018}. This analog approach means that reservoir computers can potentially be very fast, and yet consume little power, while being small and light.

One obstacle to building analog reservoir computers is that creating and connecting the analog nodes may be difficult, which is why some photonic systems \cite{larger2012, van_der_sande2017} use one actual node and then use time multiplexing to create a set of virtual nodes by adding a delay loop to the laser system. This time multiplexing increases the number of nodes but slows the response time of the photonic reservoir computer. Other types of analog reservoir computers, such as those constructed from field programmable gate arrays \cite{canaday2018} have nodes that are coupled directly, but in most systems the coupling of large numbers of nonlinear nodes into a reservoir computer network is still a research problem.

It has been shown that the number of linearly independent signals in a reservoir computer is an indicator of how well a reservoir computer can fit signals .  In \cite{carroll2019, carroll2020} this number was measured as the covariance rank. A related measure of capacity was proposed in \cite{dambre2012}. The covariance rank of a reservoir computer is ultimately limited by the number of nodes in the network, but in some cases the rank is less than the number of nodes.

In this work I show that it is possible to increase the covariance rank of a reservoir computer by adding filters to the reservoir computer output. I also note that adding filters to a reservoir computer can increase the memory capacity of a reservoir computer without affecting the nonlinearity. There are many types of filters that could be used- to keep things simple, in this work I use linear finite impulse response (FIR) filters. Infinite impulse response (IIR) filters could also be used, but they can increase the fractal dimension of a signal \cite{badii1988}, so FIR filters are used to avoid this complication. Filter design and implementation is a mature technology, and filters can be implemented in off the shelf devices such as field programmable gate arrays (FPGAs) \cite{fpga2020}.

I begin by describing how adding functions to a reservoir computer may increase the rank of the reservoir computer (Section \ref{filtrank}). Section \ref{computers} then describes reservoir computers and how filters may be added to increase their rank.  Section \ref{fitting} shows how adding filters improves signal fitting, Section \ref{ppredict} describes the impact on prediction, and signal classification is discussed in Section \ref{class}. Section \ref{mem_cal} describes changes in the memory capacity caused by adding filters.

\section{Filters and Rank}
\label{filtrank}
Given a signal $x(n)$ and some function $f(x)$, one may create a basis of rank 2. Applying Gram-Schmidt orthogonalization,
\begin{equation}
\label{gs}
\begin{array}{*{20}{l}}
{{u_1}\left( n \right) = \frac{{x\left( n \right)}}{{\left\| {x\left( n \right)} \right\|}}}\\
{y\left( n \right) = \frac{{f\left( x \right)}}{{\left\| {f\left( x \right)} \right\|}}}\\
{z\left( n \right) = y\left( n \right) - \left\langle {{u_1}\left( n \right),y\left( n \right)} \right\rangle {u_1}\left( n \right)}\\
{{u_2}\left( n \right) = \frac{{z\left( n \right)}}{{\left\| {z\left( n \right)} \right\|}}}
\end{array}
\end{equation}
where $\left\langle {} \right\rangle $ indicates a dot product and $\left\| x \right\| = \left\langle {x,x} \right\rangle $. The signals $u_1(n)$ and $u_2(n)$ form an orthonormal basis of rank 2, as long as $f(x)$ is not just a multiple of $x$. The Gram-Schmidt procedure may be repeated to create bases of higher rank using additional functions.

One type of function that can be used for $f$ is a linear filter, in particular a finite impulse response (FIR) filter. An FIR filter is a linear system with no feedback. The design of FIR filters uses well established techniques, and implementing FIR filters in hardware is also well known; a field programable gate array (FPGA) may be used to implement these filters, for example. Because FIR filters have no feedback, stability is not a question. It is possible to design stable infinite response filters (which include feedabck), but it is known that filters with feedback can increase the fractal dimension of a signal from a chaotic dynamical system  \cite{badii1988}, a complication that is avoided with FIR filters.

\subsection{FIR filters}
\label{firfilt}
An FIR filter may be described by
\begin{equation}
\label{fir}
y\left( t \right) = \sum\limits_{k = 0}^{{N_F}} {{a_k}x\left( {t - k} \right)} 
\end{equation}
where $N_F$ is the filter order.

The particular type of FIR filter I used was a Bessel filter \cite{tietze1991}. The denominator of the transfer function of an $n'th$ order Bessel filter is the $n'th$ order Bessel polynomial (the numerator is a constant). Other filters may also work, and multiple different types of filters could be used simultaneously- low order Bessel filters were used here because they are simple to describe.

I used Bessel filters with orders from 1 to 5. The Bessel filters were designated
\begin{equation}
\label{firbank}
y_i^\eta \left( t \right) = \sum\limits_{k = 1}^\eta  {{a_k}{\chi_i}\left( {t - k} \right)} 
\end{equation}
where $i$ indicates the node index from the reservoir computer and $\eta$ is the filter order. The filter coefficients are given in Table \ref{filtcoeff}.

\begin{table}[]
\centering
\caption{Filter coefficients}
\label{filtcoeff}
\begin{tabular}{|c|c|c|c|c|c|}
\hline
 filter number & $k=1$ &  $k=2$ &  $k=3$  & $k=4$  & $k=5$  \\
\hline
 1 & 1 & 0 & 0 & 0 & 0 \\
 2 & 1.7321 & 1 & 0 & 0 & 0 \\
3 & 2.4329 & 2.4662 & 1 & 0 & 0 \\
 4 & 3.1239 & 4.3916 & 3.2011 & 1 &  0\\
 5 & 3.8107 & 6.7767 & 6.8864 & 3.9363 & 1 \\

\hline         
\end{tabular}
\end{table}

\section{Reservoir Computers}
\label{computers}

Two different node types are used for the reservoir computers in this work. The first reservoir computer uses leaky tanh nodes \cite{jaeger2007}, 
\begin{equation}
\label{umd_comp}
{\chi _i}\left( {n + 1} \right) = \alpha {\kern 1pt} {\chi _i}\left( n \right) + \left( {1 - \alpha } \right)\tanh \left( {\sum\limits_{j = 1}^M {{A_{ij}}{\chi _j}\left( n \right)}  + {w_i}s\left( n \right) + 1} \right)
\end{equation}
where the reservoir computer variables are $\chi_i(n), i=1 ... M$ with $M$ the number of nodes, $A$ is an adjacency matrix that described how the different nodes in the network are connected to each other, ${\bf W}=[w_1, w_2, ... w_M]$ describes how the input signal $s(n)$ is coupled into the different nodes, and $f$ is a nonlinear function. 
 
 For the leaky tanh map reservoir computer, half of the elements of the adjacency matrix $A$ were chosen randomly and  set to random numbers drawn from a uniform random distribution between -1 and 1. The diagonal elements of $A$ were then set to zero. The spectral radius $\sigma$ is the largest magnitude of the eigenvectors of $A$. The entire adjacency matrix was renormalized to have a spectral radius specified for the different examples below.

The second node type was a model for the laser experiment of \cite{larger2012}. This system is described by
\begin{equation}
\label{mzeqn}
\varepsilon \dot x\left( s \right) + x\left( s \right) = \beta {\sin ^2}\left[ {\mu x\left( {s - 1} \right) + \rho {u_I}\left( {s - 1} \right) + \phi } \right]
\end{equation}
where $s$ is a normalized time.

Converting $s$ and $x$ to discrete variables, this system may be modeled by the map
\begin{equation}
\label{mzmap}
x\left( {n + {\tau _s}} \right) = \sum\limits_{j = 1}^N {\beta H\left( j \right){{\sin }^2}\left[ {\mu x\left( {n - j} \right) + {\bf{W}}s\left( {\left\lfloor {\frac{n}{M}} \right\rfloor } \right) + \phi } \right]} 
\end{equation}
where $\left\lfloor {} \right\rfloor $ is the floor function and $M$ is the number of nodes. The floor function means that the value of the input signal is sampled once every $M$ time steps of the map. The variable $\tau_s$ in this equation is an integer. The variable $\beta$ was set at 0.5, $\mu=0.1$ and $\phi=0$. The signal $H(j)$ is the impulse response of the low pass filter in eq. (\ref{mzeqn}). The low pass filter was a first order filter with a time constant of $\tau_R=1.5 \times 10^{-6}$. The elements of the vector ${\bf W}$ are drawn from a uniform random distribution between +1 and -1. 

The time step in the map of eq. (\ref{mzmap}) was $t_s=7.5 \times 10^{-8}$ s. The number of nodes $M$ was varied, so the total time for one update of the reservoir computer was $M \times t_s$.

The map signal $x(n)$ was rearranged into a matrix $\Omega$ to create the set of reservoir computer nodes. 
\begin{equation}
\label{network}
\begin{array}{*{20}{l}}
{n = 1,2 \ldots N}\\
{i = \left( {n\;\,\bmod \,\;M} \right) + 1}\\
{j = \left\lfloor {\frac{n}{M}} \right\rfloor  + 1}\\
{{\Omega _{i,j}} = x\left( n \right)}
\end{array}
\end{equation}

Each column of this matrix represented one virtual node, while each row corresponded to one time step for the input signal.  The parameters for eq. (\ref{network}) were taken from the experiment in \cite{larger2012}.

\subsection{Training and Testing}
When the reservoir computer was driven with $s(n)$, the first 1000 time steps were discarded as a transient. The next $N=10000$ time steps from each node were combined in a $N \times (M+1)$ matrix
\begin{equation}
\label{fit_mat}
\Omega  = \left[ {\begin{array}{*{20}{c}}
{{\chi _1}\left( 1 \right)}& \ldots &{{\chi _M}\left( 1 \right)}&1\\
{{\chi _1}\left( 2 \right)}&{}&{{\chi _M}\left( 2 \right)}&1\\
 \vdots &{}& \vdots & \vdots \\
{{\chi _1}\left( N \right)}& \ldots &{{\chi _M}\left( N \right)}&1
\end{array}} \right]
\end{equation}
The last column of $\Omega $ was set to 1 to account for any constant offset in the fit. The training signal is fit by

\begin{equation}
\label{train_fit}
{h(n)} ={\Omega } {{\bf C}}
\end{equation}
where ${h(n)} = \left[ {h\left( 1 \right),h\left( 2 \right) \ldots h\left( N \right)} \right]$ is the fit to the training signal ${g(n)} = \left[ {g\left( 1 \right),g\left( 2 \right) \ldots g\left( N \right)} \right]$ and ${{\bf C}} = \left[ {{c_1},{c_2} \ldots {c_N}} \right]$ is the coefficient vector. The fit coefficient vector ${\bf C}$ is found by ridge regression.

The training error may be computed from
\begin{equation}
\label{train_err}
{\Delta _{RC}} = \frac{{{\rm{std}}\left[ {\Omega {\bf{C}} - g(n)} \right]}}{{{\rm{std}}\left[ {g(n)} \right]}}
\end{equation}
where std[ ] indicates a standard deviation. 

In the testing configuration, the reservoir computer is driven by a new signal $s'(n)$ and the the matrix of signals from the reservoir is now $\Omega'$.  The testing error is
\begin{equation}
\label{test_err} 
{\Delta _{tx}} = \frac{{{\rm{std}}\left[ {\Omega '{\bf{C}} - g'\left( n \right)} \right]}}{{{\rm{std}}\left[ {g'\left( n \right)} \right]}}
\end{equation}
where $g'(n)$ is the testing signal.

\subsection{Adding Filters}
\label{addfilt}

Each node output $\chi_i(n)$ was passed through between one and five filters, with filter coefficients defined in Table \ref{filtcoeff}. The filter outputs were $y_i^\eta \left( t \right)$, found as
\begin{equation}
\label{firbank}
y_i^\eta \left( t \right) = \sum\limits_{k = 1}^\eta  {{a_k}{\chi_i}\left( {t - k} \right)} 
\end{equation}
where $i$ indicates the node index from the reservoir computer and $\eta$ is the filter order. 

The filter outputs were arranged in a matrix similar in form to $\Omega$
\begin{equation}
\label{filtmat}
\Lambda  = \left[ {\begin{array}{*{20}{c}}
{y_1^1\left( 1 \right)}&{y_1^2\left( 1 \right)}& \ldots &{y_2^1\left( 1 \right)}& \ldots &{y_M^{{N_F}}\left( 1 \right)} & 1\\
{y_1^1\left( 2 \right)}&{}&{}&{}&{}& \vdots & \vdots \\
 \vdots &{}&{}&{}&{}& \vdots & \vdots\\
{y_1^1\left( N \right)}&{y_1^2\left( N \right)}& \ldots &{y_2^1\left( N \right)}& \ldots &{y_M^{{N_F}}\left( N \right)} & 1
\end{array}} \right]
\end{equation}

For $N$ time series points, $M$ nodes and a filter order of $N_F$, the size of $\Lambda$ is $N \times (M \times N_F + 1)$.

Fitting proceeds as with the reservoir only, but using the full $\Lambda$ matrix instead; the fit coefficients are found as 
\begin{equation}
\label{fit_coeff_filt}
{{\bf C}_F} = {{ \Lambda } _{inv}}{g(n)}
\end{equation}
where $g(n)$ is the same training signal as was used for the reservoir. The training and testing errors are calculated in the same manner as for the reservoir, substituting $\Lambda$ for $\Omega$.

\subsection{Covariance Rank}
\label{comp_rank}
The individual columns of the reservoir matrix $\Omega$ or the filter matrix $\Lambda$ may be used as a basis to fit the training signal $g(n)$. Among other things, the fit will depend on the number of orthogonal columns in $\Omega$.

Principal component analysis \cite{joliffe2011} states that the eigenvectors of the covariance matrix of $\Omega$, $\Theta=\Omega^T\Omega$, form an uncorrelated basis set. The rank of the covariance matrix tells us the number of uncorrelated vectors. 

Therefore, we will use the rank of the covariance matrix of $\Omega$,
\begin{equation}
\label{rank}
\begin{array}{l}
\Gamma \left( \Omega  \right) = {\rm{rank}}\left( {{\Omega ^T}\Omega } \right)\\
\Gamma \left( \Lambda  \right) = {\rm{rank}}\left( {{\Lambda ^T}\Lambda } \right)
\end{array}
\end{equation}
to characterize the reservoir matrix $\Omega$ or the filter matrix $\Lambda$. We calculate the rank using the MATLAB rank() function. The maximum covariance rank is equal to the number of nodes, $M$ or the number of nodes times the number of filters. In \cite{carroll2019, carroll2020}, higher covariance rank was associated with lower testing error.

\section{Signal Fitting}
\label{fitting}
The first example of adding filters to a reservoir computer will be fitting the $z$ signal from the Lorenz chaotic system based on the $x$ signal. the The Lorenz system \cite{lorenz1963} is described by
\begin{equation}
\label{lorenz}
\begin{array}{l}
\frac{{dx}}{{dt}} = {c_1}y - {c_1}x\\
\frac{{dy}}{{dt}} = x\left( {{c_2} - z} \right) - y\\
\frac{{dz}}{{dt}} = xy - {c_3}z
\end{array}
\end{equation}
with $c_1$=10, $c_2$=28, and $c_3$=8/3. The equations were numerically integrated with a time step of $t_s=0.02$.

\subsection{Leaky Tanh Nodes}

\begin{figure}
\centering
\includegraphics[scale=0.8]{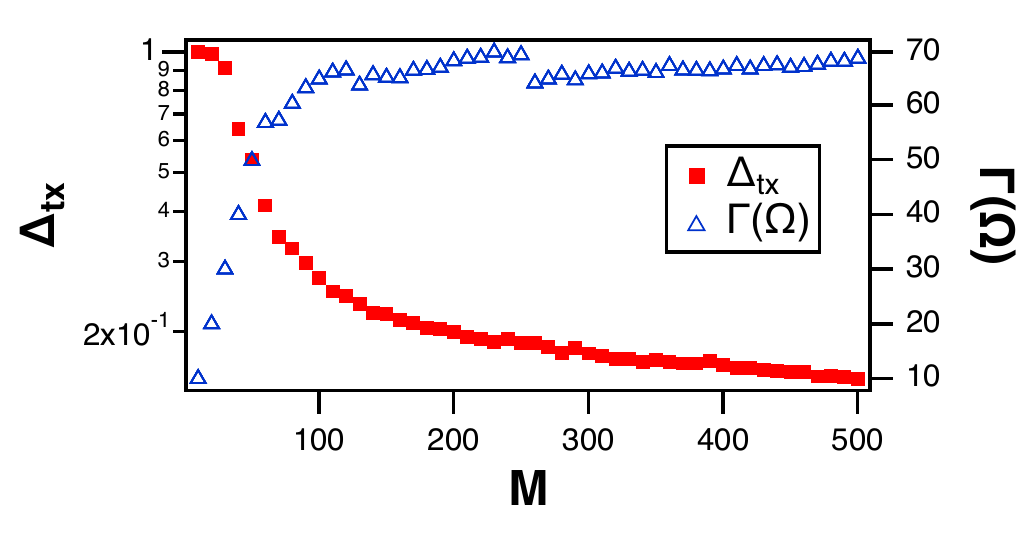} 
  \caption{ \label{umdfits} Testing error $\Delta_{tx}(\Omega)$ and the covariance rank $\Gamma(\Omega)$ as a function of the number of nodes $M$ for the leaky tanh nodes of eq. (\ref{umd_comp}) when the input signal is the Lorenz $x$ signal and the testing signal is the Lorenz $z$ signal. These quantities were calculated from the reservoir signal matrix as in eq (\ref{fit_mat}). The reservoir computer is based on the leaky tanh nodes of eq. (\ref{umd_comp}).}
  \end{figure}

The optimum parameters for fitting the Lorenz $z$ signal when the Lorenz $x$ signal drove the leaky tanh reservoir computer were found by simulations to be $\alpha=0.75$ and a spectral radius of 0.48. Figure \ref{umdfits} shows the testing error $\Delta_{tx}(\Omega)$ and the covariance rank $\Gamma(\Omega)$ as a function of the number of nodes $M$ for the leaky tanh nodes. The notation $(\Omega)$ is used to indicate that the testing error and the covariance rank are calculated from the reservoir signal matrix $\Omega$ defined in eq. \ref{fit_mat}.

\begin{figure}
\centering
\includegraphics[scale=0.8]{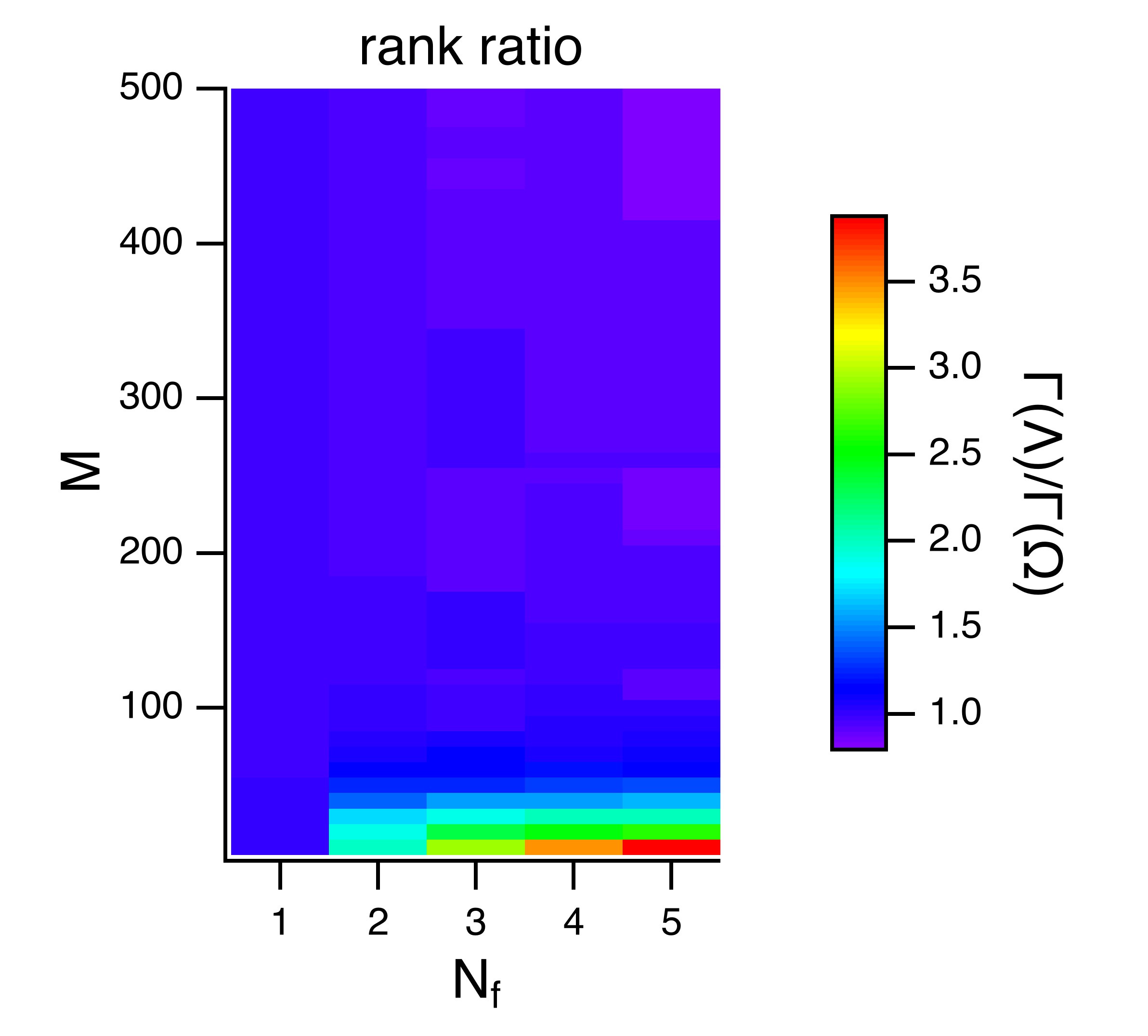} 
  \caption{ \label{umdrankratio} Ratio of covariance rank $\Gamma(\Lambda)$ when $N_f$ filters are added after the reservoir to the testing error found using only the reservoir, $\Gamma(\Omega)$. The number of filters used is $N_f$ while the number of nodes is $M$. The reservoir computer is based on the leaky tanh nodes of eq. (\ref{umd_comp}).}
  \end{figure}

  Figure \ref{umdfits} shows a typical result, that the covariance rank increases and the testing error decreases as the number of reservoir computer nodes goes up.   
   Figure \ref{umdrankratio} shows the ratio of the covariance rank calculated from the filter matrix $\Lambda$ of eq. (\ref{filtmat}) to the covariance rank  calculated from the reservoir matrix $\Omega$. Figure \ref{umdrankratio} shows that adding FIR filters as described in Section \ref{firfilt} can increase the covariance rank  for the leaky tanh nodes if the reservoir is small, 20 nodes or less.  For larger reservoirs, the improvement in rank is not as large. It was seen in figure \ref{umdfits} that the covariance rank of the reservoir matrix $\Omega$ saturated when the reservoir computer reached 100 nodes. Part of this saturation may be numerical in nature; the MATLAB rank algorithm sets a threshold below which singular values are considered to be zero. If the reservoir matrix $\Omega$ is too large, numerical errors may cause a number of singular values to be below this threshold. For the same reason, the lack of increase in rank of the filter matrix $\Lambda$ relative to the reservoir matrix may be caused by these round off errors.
  
 \begin{figure}
\centering
\includegraphics[scale=0.8]{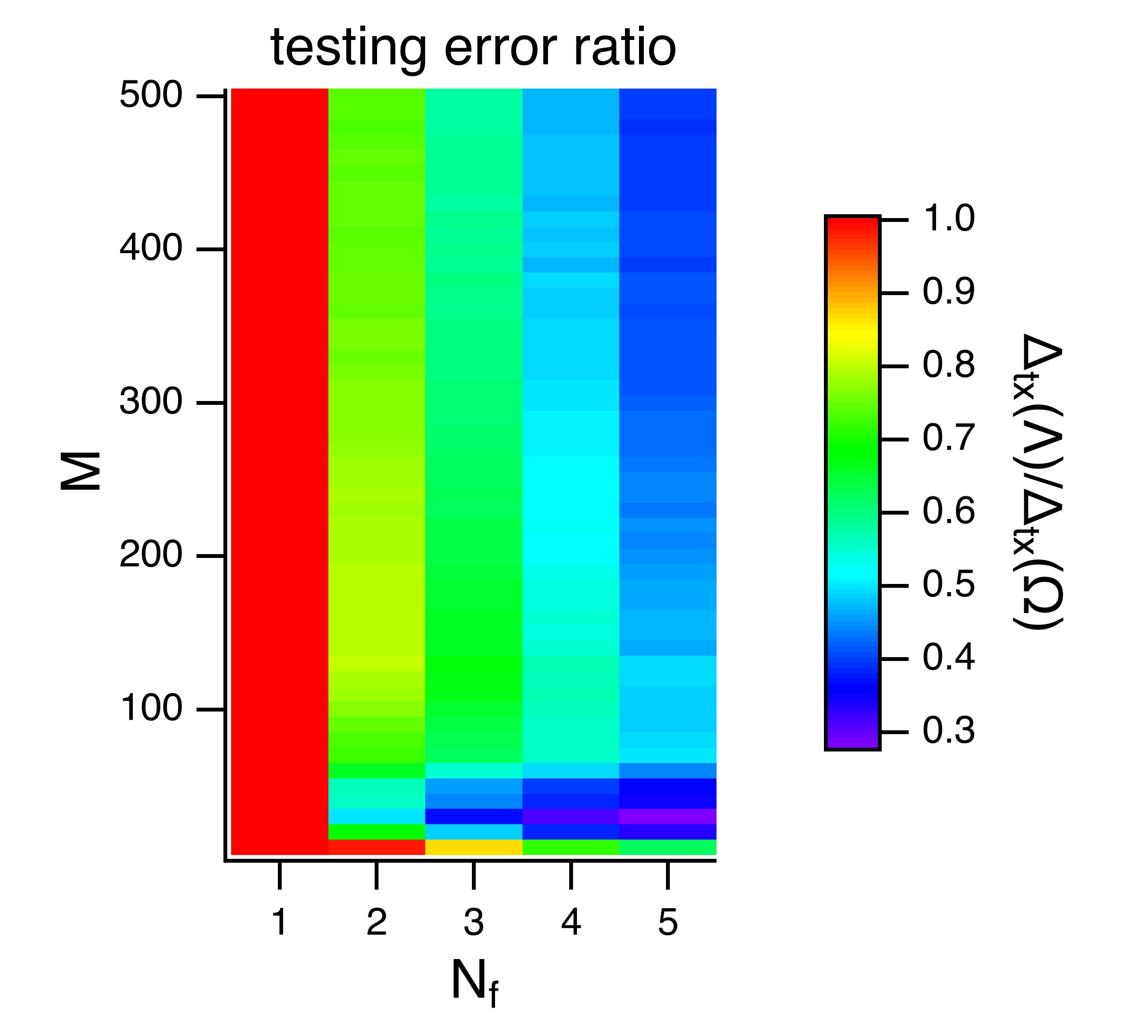} 
  \caption{ \label{umderrratio} Ratio of testing error $\Delta_{tx}(\Lambda)$ when $N_f$ filters are used after the reservoir to the testing error found using only the reservoir, $\Delta_{tx}(\Omega)$. The number of filters used is $N_f$ while the number of nodes is $M$. The reservoir computer is based on the leaky tanh nodes of eq. (\ref{umd_comp}).}
  \end{figure}   

Figure \ref{umderrratio} shows the ratio of the testing error calculated from the filter matrix $\Lambda$ of eq. (\ref{filtmat}) to the testing error  calculated from the reservoir matrix $\Omega$ of eq. (\ref{fit_mat}).  When a single filter is used, it is a first order Bessel filter, and Table \ref{filtcoeff} shows that the first order Bessel filter is an identity, so the testing error for one added filter is the same as the testing error for the reservoir computer by itself. For reservoir computers of 20 nodes or less, adding filters leads to large improvements in the testing error. For larger reservoir computers, the testing error is still improved when filters are added, but not by as much. The fact that testing errors still drop relative to the testing errors for the reservoir without filters suggests that the saturation of covariance rank in figures \ref{umdfits} and \ref{umdrankratio} is caused by numerical effects.

The effect of the filters on the testing error may depend on the particular node parameters used, so a set of randomly defined reservoir computers with 80 nodes was simulated to look for variations as a function of parameter. The parameter $\alpha$ in eq. (\ref{umd_comp}) was varied from 0.05 to 1, while the spectral radius $\sigma$ was varied from 0.1 to 3. For each value of $\alpha$ and $\sigma$, 20 random adjacency matrices and 20 random input vectors ${\bf W}$ were generated. The elements of ${\bf W}$ were chosen from a uniform random distribution between -1 and 1, while a random selection of half of the elements of the adjacency matrix $A$ were set to zeroThe nonzero elements of $A$ were set to uniformly distributed random numbers between -1 and 1. The diagonal elements of $A$ were then set to zero.

\begin{figure}[h]
\centering
\includegraphics[scale=0.8]{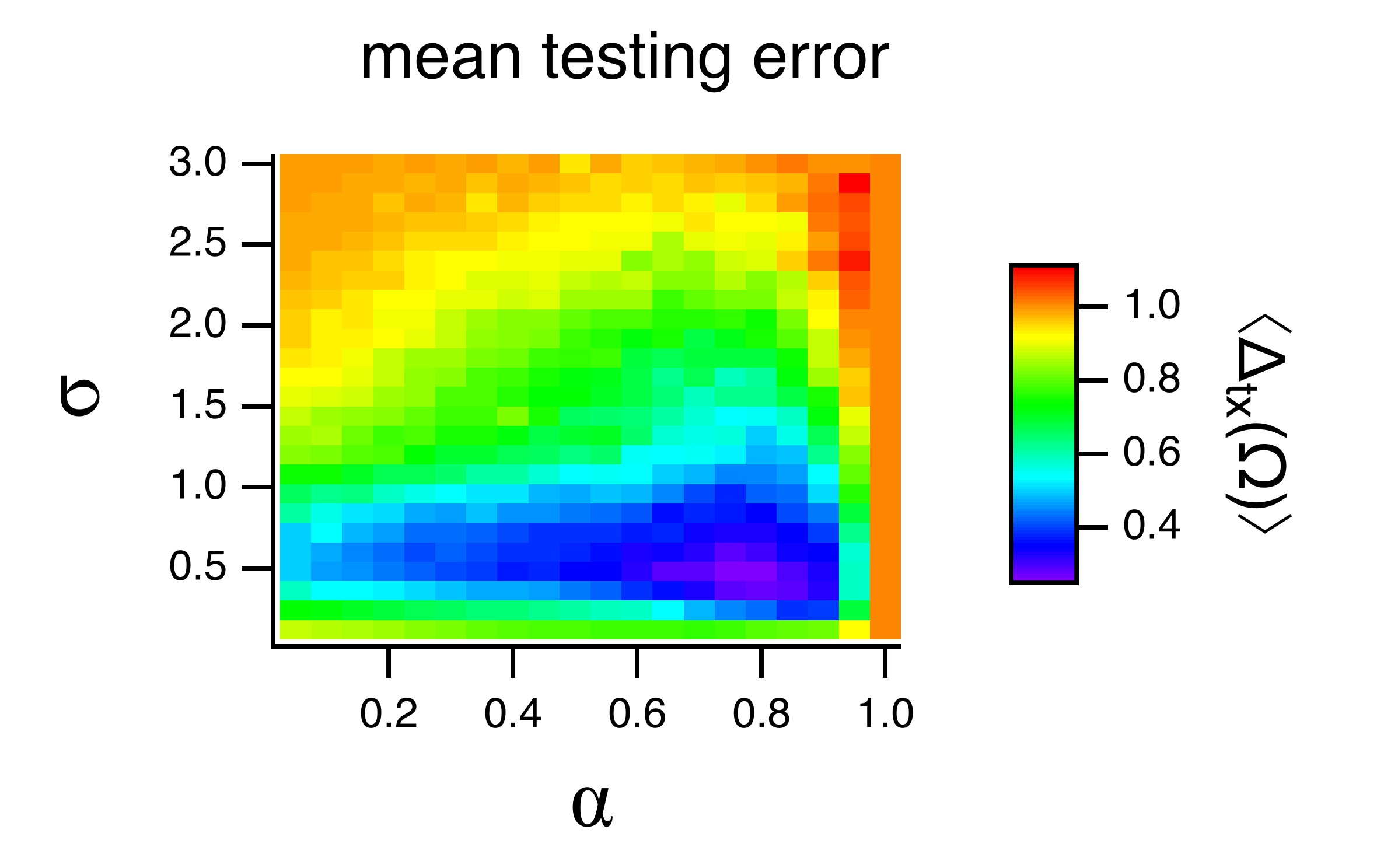} 
  \caption{ \label{umd_err_mean} Mean testing error $\left\langle {{\Delta _{tx}}(\Omega )} \right\rangle$ as a function of the parameter $\alpha$ and the spectral radius $\sigma$ for a leaky tanh reservoir with 80 nodes.}
  \end{figure}

Figure \ref{umd_err_mean} shows the mean testing error for the leaky tanh reservoir computer as the parameter $\alpha$ and the spectral radius $\sigma$ are chenged.

The mean testing error is smallest near $\sigma=0.5$ and $\alpha=0.8$.

\begin{figure}[h]
\centering
\includegraphics[scale=0.8]{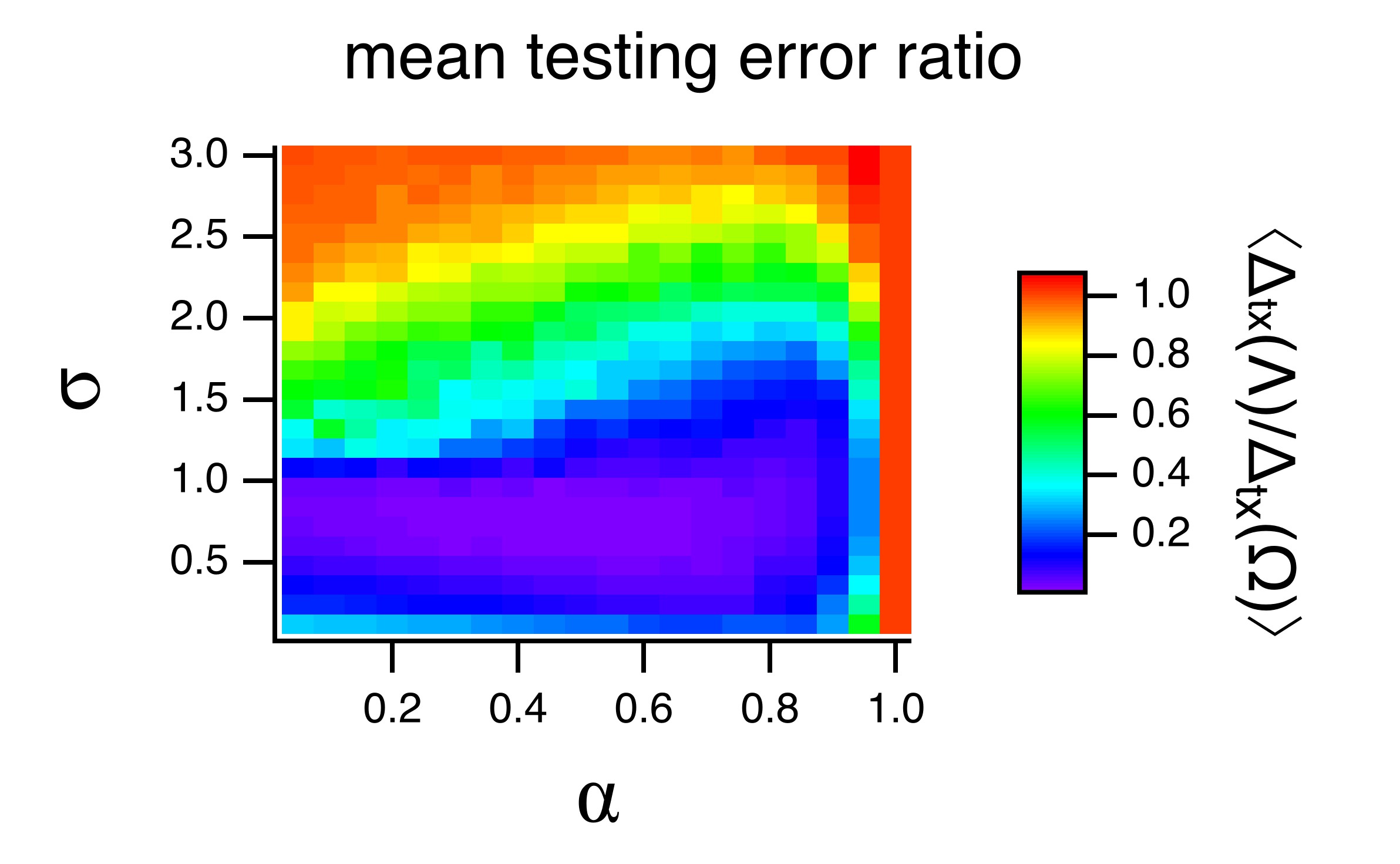} 
  \caption{ \label{umd_mean_err_ratio} Mean ratio of the testing error $\Delta_{tx}(\Lambda)$ when five filters are used in the filter matrix $\Lambda$ to the testing error for the leaky tanh reservoir by itself, as a function of the parameter $\alpha$ and the spectral radius $\sigma$. The reservoir had 80 nodes}
  \end{figure}

Figure \ref{umd_mean_err_ratio} shows the mean ratio of the testing error when five filters are used for the filter matrix $\Lambda$ to the testing error from the reservoir only, or $\left\langle {{\Delta _{tx}}\left( \Lambda  \right)/{\Delta _{tx}}\left( \Omega  \right)} \right\rangle $. 

  Figures \ref{umd_err_mean} and \ref{umd_mean_err_ratio} show that the largest improvement in testing error caused by adding filters comes in the parameter region where the reservoir computer by itself already has the smallest testing error. Possibly the regions of poor fit in both figures occur because the reservoir signals are not a good match for the training signal, so increasing the rank of the reservoir computer does not provide any benefit. For $\alpha=1$ in figures \ref{umd_err_mean} and \ref{umd_mean_err_ratio}, $\chi_i(n+1) = \chi_i(n)$ in eq. (\ref{umd_comp}), so the reservoir computer does not respond to the input signal.

\subsection{Laser system}

 \begin{figure}[h]
\centering
\includegraphics[scale=0.8]{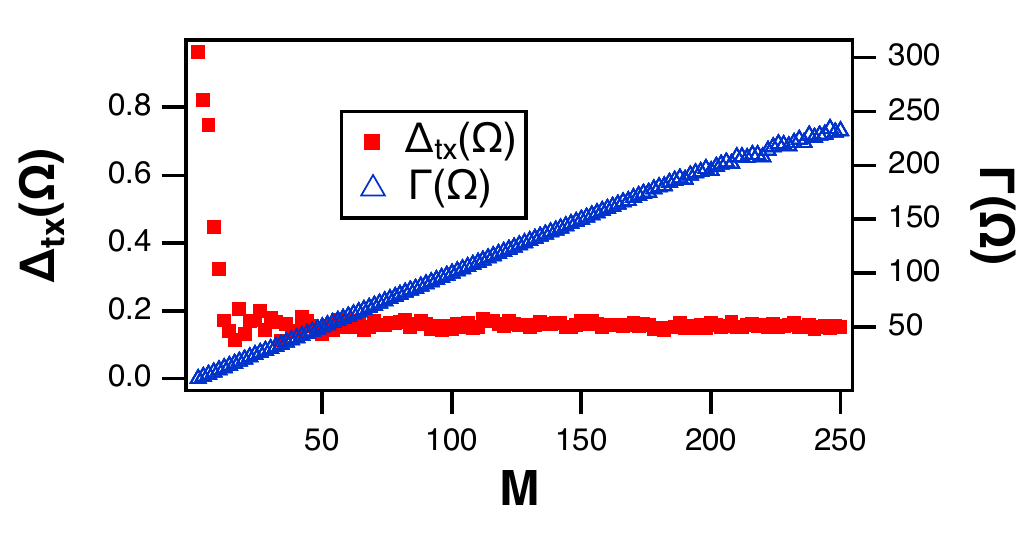} 
  \caption{ \label{laserfits} Testing error $\Delta_{tx}(\Omega)$ and the covariance rank $\Gamma(\Omega)$ as a function of the number of nodes $M$ for the laser system nodes of eqs. (\ref{mzmap}-\ref{network}) when the input signal is the Lorenz $x$ signal and the testing signal is the Lorenz $z$ signal. These quantities were calculated from the reservoir signal matrix as in eq (\ref{fit_mat}).}
  \end{figure} 
  
The leaky tanh nodes of the previous section are usually implemented on a digital computer, in which case  there is no advantage to adding filters. The laser system of \cite{larger2012} was built as an actual experiment. The performance of the reservoir computer could be improved by adding more virtual nodes, but adding virtual nodes slowed the response time, so there is a more obvious advantage to adding filters to this system. 

For a filter of order $\eta$, there is a startup delay in the FPGA of $\eta/2$. After that the filter runs in real time; one filter time step equals one laser system time step. For $M$ nodes, the time for one full update of the reservoir computer will be $M \times t_s$. If instead we have $M_f$ nodes followed by filters, the update time is $M_f \times t_s \times (1+\eta/2)$, where $\eta$ is the maximum filter order. We gain in speed if $M_f < M/(1+\eta/2)$. For five filters with a maximum order of $\eta=5$, $M/(1+\eta/2)=0.28M$, while $M_f=0.2M,$ so there is a speedup factor of 1.4

Up to the capacity of the FPGA, the filters can be operated in parallel, so by using different types of filters, or filters with different frequency or phase characteristics, we could combine more than $\eta$ filters with a maximum order of $\eta$ to achieve even a greater speedup. It is also possible to use both FIR and IIR (infinite impulse response) filters, as long as the IIR filters are stable.

The map of eqs. (\ref{mzmap}-\ref{network}) was used to simulate the laser reservoir computer. The parameters for the laser map were taken from the experiment in \cite{larger2012}.

   \begin{figure}[h]
\centering
\includegraphics[scale=0.8]{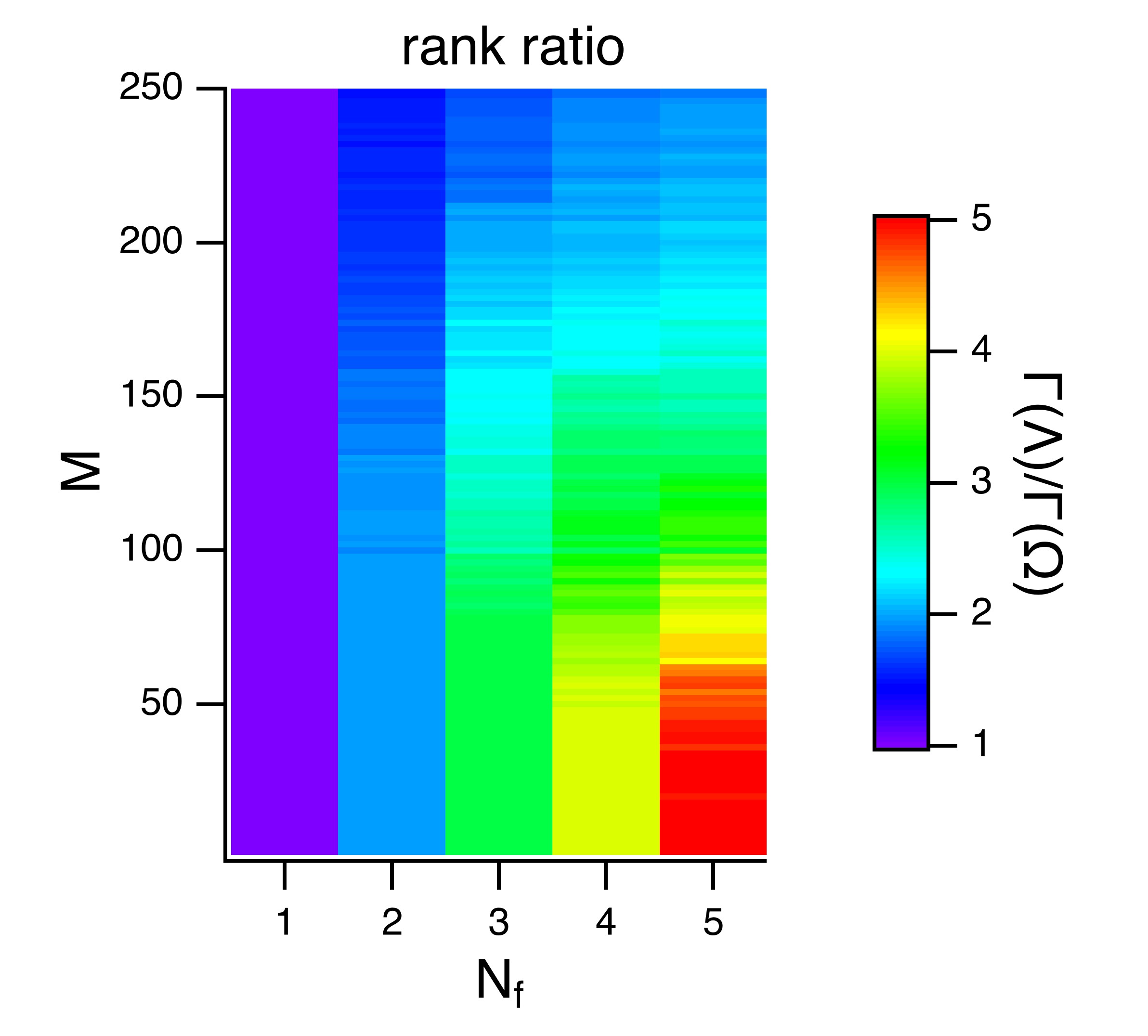} 
  \caption{ \label{laserrankratio} Ratio of covariance rank $\Gamma(\Lambda)$ when $N_f$ filters are used after the reservoir to the testing error found using only the reservoir, $\Gamma(\Omega)$. The number of filters used is $N_f$ while the number of nodes is $M$. The reservoir computer was modeled by the laser system of eqs. (\ref{mzmap}-\ref{network}).}
  \end{figure} 

Figure \ref{laserfits} shows the testing error and covariance rank for the laser simulation of eqs. (\ref{mzmap}-\ref{network}). The testing error stops decreasing as the number of nodes becomes greater than $M=20$. Instead of an adjacency matrix, the coupling between nodes in the laser system comes from the low pass filter that is part of the delay loop. The low pass filter creates a fading memory in the laser reservoir computer, but it also means that nodes separated by a number of time steps longer than the memory of the low pass filter are not coupled. As a result, increasing the reservoir past a certain size does not lead to a further decrease in testing error. The ratio of the low pass filter time constant to the reservoir time step was $\tau_R/t_s=20$, which is about the number of nodes for which the testing error stops decreasing.

     \begin{figure}[h]
\centering
\includegraphics[scale=0.8]{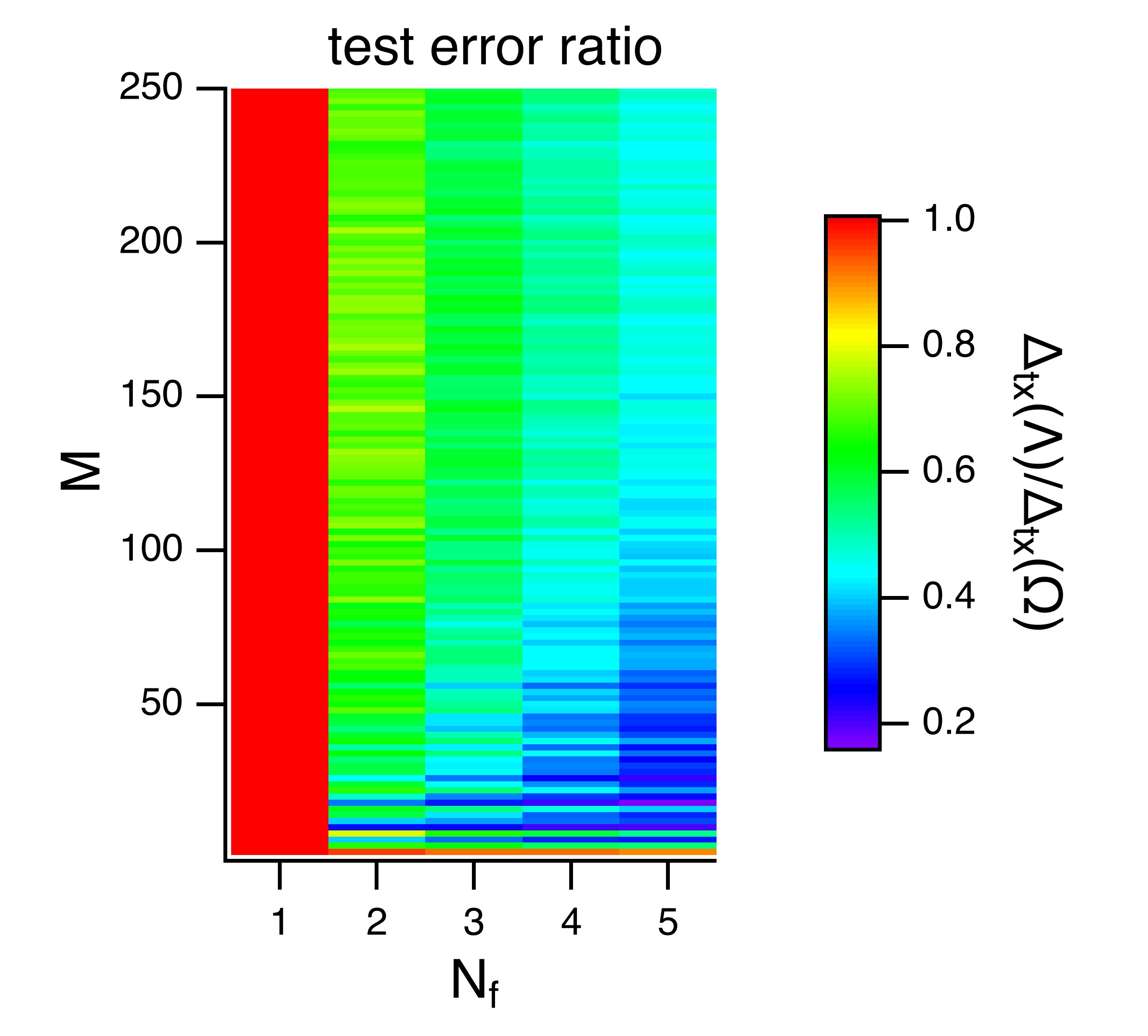} 
  \caption{ \label{lasererrratio} Ratio of testing error $\Delta_{tx}(\Lambda)$ when $N_f$ filters are used after the reservoir to the testing error found using only the reservoir, $\Delta_{tx}(\Omega)$. The number of filters used is $N_f$ while the number of nodes is $M$. The reservoir computer was modeled by the laser system of eq. (\ref{mzmap}-\ref{network}).}
  \end{figure} 
  
  Figure \ref{laserrankratio} shows the ratio of the covariance rank calculated from the filter matrix $\Lambda$ of eq. (\ref{filtmat}) to the covariance rank  calculated from the reservoir matrix  $\Omega$ of eq. (\ref{fit_mat}) for the laser reservoir computer simulation. As in figures \ref{umdrankratio} and \ref{umderrratio}, when only a single filter is present, it is equivalent to the identify.
  
 For reservoirs up to about 50 nodes, adding filters increased the rank of the filter matrix $\Lambda$ relative to the reservoir matrix $\Omega$ in proportion to the number of added filters. As the reservoir computer became larger, the increase in rank was not as large. Figure \ref{laserfits} shows that the covariance rank stops increasing with the number of nodes for a reservoir computer with about 250 nodes. When five filters are added to the reservoir computer, the filter matrix $\Lambda$ will have dimensions $250 \times 251$. It is possible that numerical errors limit the rank calculation for matrices of this size or larger. 
  
 The improvement in testing error when the laser reservoir computer is followed by a set of filters is shown in figure \ref{lasererrratio}. This plot is similar to figure \ref{umderrratio} for the leaky tanh nodes: the greatest improvement in testing error comes for a small number of nodes. 
 
\begin{figure}[h]
\centering
\includegraphics[scale=0.8]{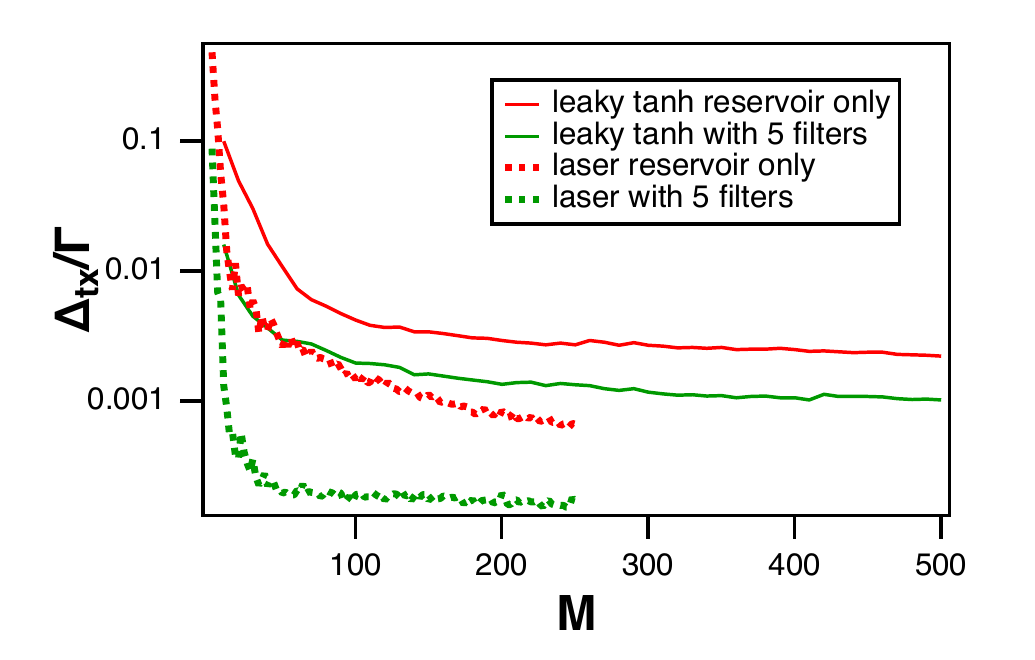} 
  \caption{ \label{errtorank} Ratio of testing error $\Delta_{tx}$ to covariance rank $\Gamma$ for both types of reservoir computer, for the reservoir computer only and the reservoir computer followed by five filters.}
  \end{figure}

Adding filters to a reservoir computer does result in lower testing error for the same covariance rank, so there is more to adding filters than just increasing the rank of the reservoir computer. Figure \ref{errtorank} shows the ratio of testing error to covariance rank for both types of reservoir computer as the number of nodes varies.

In figure \ref{errtorank}, the ratio of testing error to covariance rank is lower for both reservoirs when five filters are added than for either reservoir by itself. Adding filters to increase the rank of the reservoir computer does lower the testing error, but the difference in testing error is greater than be accounted for by the rank alone. Adding filters also changes the memory capacity of the reservoir; these changes will be adressed in section \ref{mem_cal}.

 This section on signal fitting shows that augmenting a reservoir computer can improve performance, but the improvement is larger for reservoir computers with smaller numbers of nodes. This is not necessarily bad; if it is easy to build the reservoir computer with large numbers of nodes, then performance improvement is not as useful.

  \section{Prediction}
  \label{ppredict}
  Predicting the future time evolution of a signal given its past time evolution is a variation on fitting signals. In this case, if the input signal is $s(n)$, the training signal is $g(n)=s(n+\tau)$, where $\tau$ represents some number of time steps into the future. A useful time scale for predicting the Lorenz $x$ signal is the Lyapunov time, or the reciprocal of the largest Lyapunov exponent. For the parameters in eq. (\ref{lorenz}), the largest Lyapunov exponent is 0.9/s, so the Lyapunov time is $T_L=1.1$ s. For this section, the reservoir computers will be driven with the Lorenz $x$ signal and predict the $x$ signal $0.25T_L$ s into the future. At an integration time step of 0.02 s, this amounts to 13 points into the future. The error in prediction is calculated by a process analogous to the testing error in eq. (\ref{test_err}), but to avoid confusion the prediction error will be called $\Delta_P$.
  
 The prediction time here does not look very large, but it is a prediction time for an open loop configuration, where the output of the reservoir computer is not fed back into the input. The time interval in Lyapunov times is similar to the prediction time in \cite{jaeger2004}, which was one of the first papers to use a reservoir computer for predicting a chaotic system.
  
  \subsection{Leaky Tanh Nodes}
\begin{figure}[h]
\centering
\includegraphics[scale=0.8]{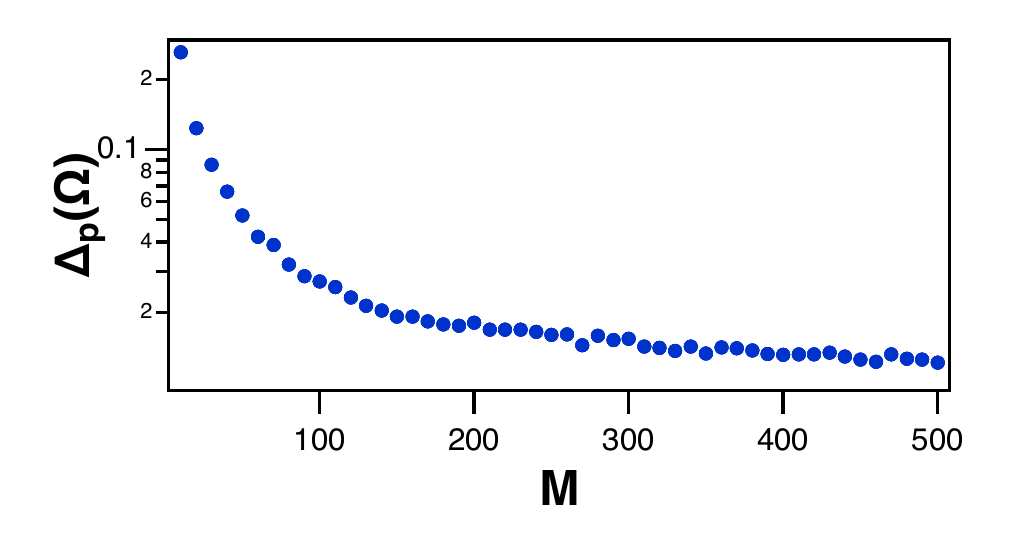} 
  \caption{ \label{pmappred} $\Delta_P(\Omega)$ is the error for predicting the future of the Lorenz $x$ signal  0.25 Lyapunov times into the future, plotted versus the number of nodes $M$. This prediction is for the reservoir computer only, using leaky tanh nodes.}
  \end{figure}

  Figure \ref{pmappred} shows the error for predicting the future of the Lorenz $x$ signal  0.25 Lyapunov times into the future, or $\Delta_P$, as a function of the number of nodes $M$ for a reservoir computer using the leaky tanh nodes (no filters).
  
\begin{figure}[h]
\centering
\includegraphics[scale=0.8]{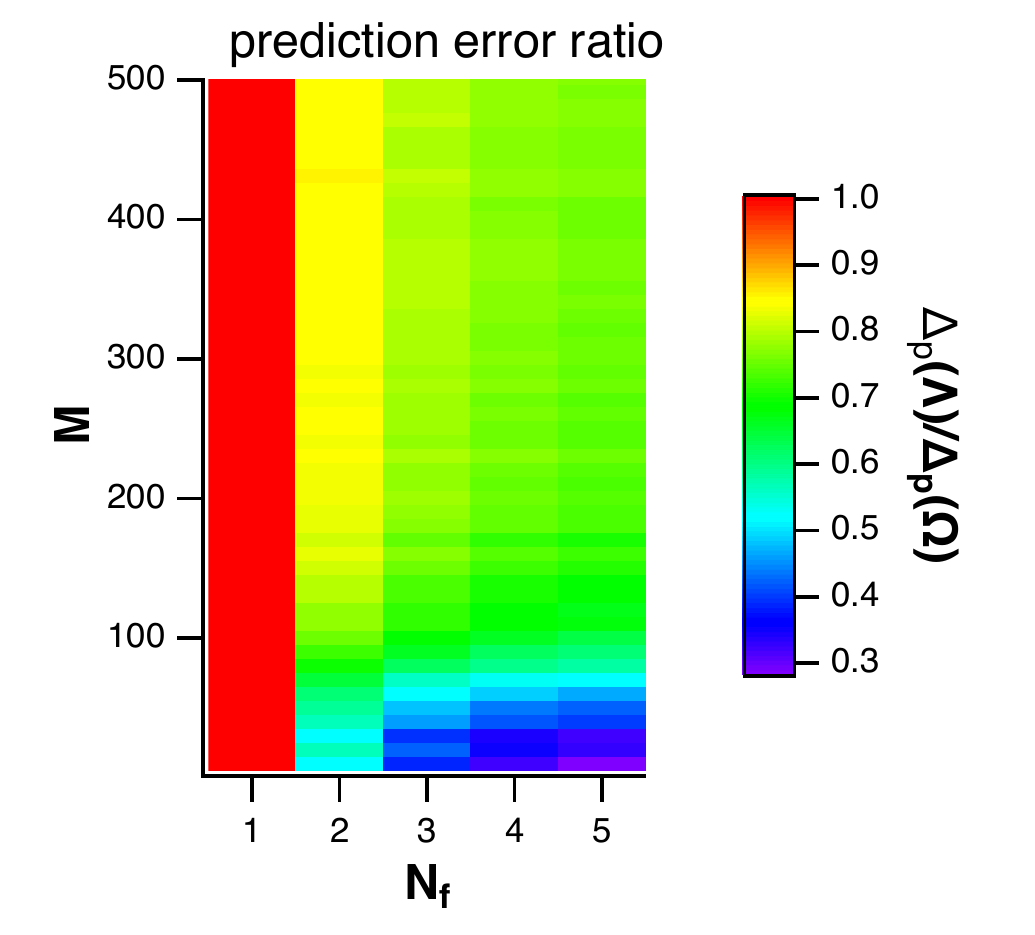} 
  \caption{ \label{pmappredfilt} Prediction error ratio $\Delta_P(\Lambda)/\Delta_P(\Omega)$ for predicting the future of the Lorenz $x$ signal when the leaky tanh reservoir computer is augmented with up to five filters. The number of nodes is $M$, while $N_f$ is the number of filters.}
  \end{figure}

 Figure \ref{pmappredfilt} shows the ratio of the prediction error using a reservoir computer augmented with filters to the prediction error using the reservoir computer only. 
 
Figure \ref{pmappredfilt} shows that adding linear filters to the leaky tanh reservoir computer can lower the error in predicting the Lorenz $x$ signal, but the improvement in prediction error is only large for reservoir computers with less than 50 nodes. 

\begin{figure}[h]
\centering
\includegraphics[scale=0.8]{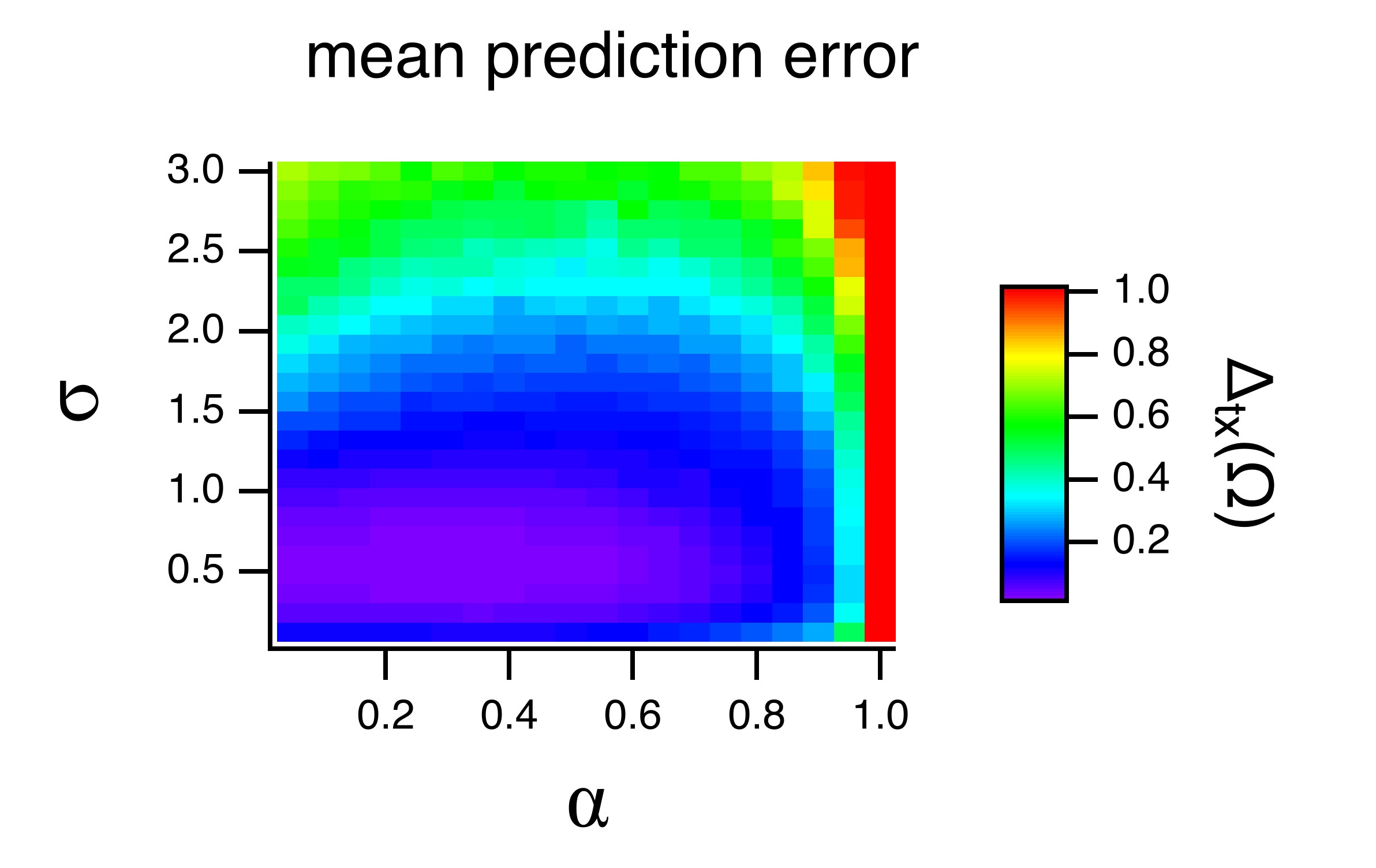} 
  \caption{ \label{umd_mean_xpred_err} Mean error in predicting the Lorenz $x$ variable,  $\Delta_{tx}(\Omega)$,  as a function of the parameter $\alpha$ and the spectral radius $\sigma$ for a leaky tanh reservoir with 80 nodes.}
  \end{figure}

The prediction error for the leaky tanh reservoir may also be evaluated for different values of the parameter $\alpha$ and the spectral radius. As with the parameter sweeps for fitting the $z$ signal, 20 random adjacency matrices were created for each value of spectral radius and $\alpha$ and the mean testing errors were plotted. Figure \ref{umd_mean_xpred_err} shows the mean error in predicting the Lorenz $x$ variable as a function of $\alpha$ and the spectral radius.

\begin{figure}[h]
\centering
\includegraphics[scale=0.8]{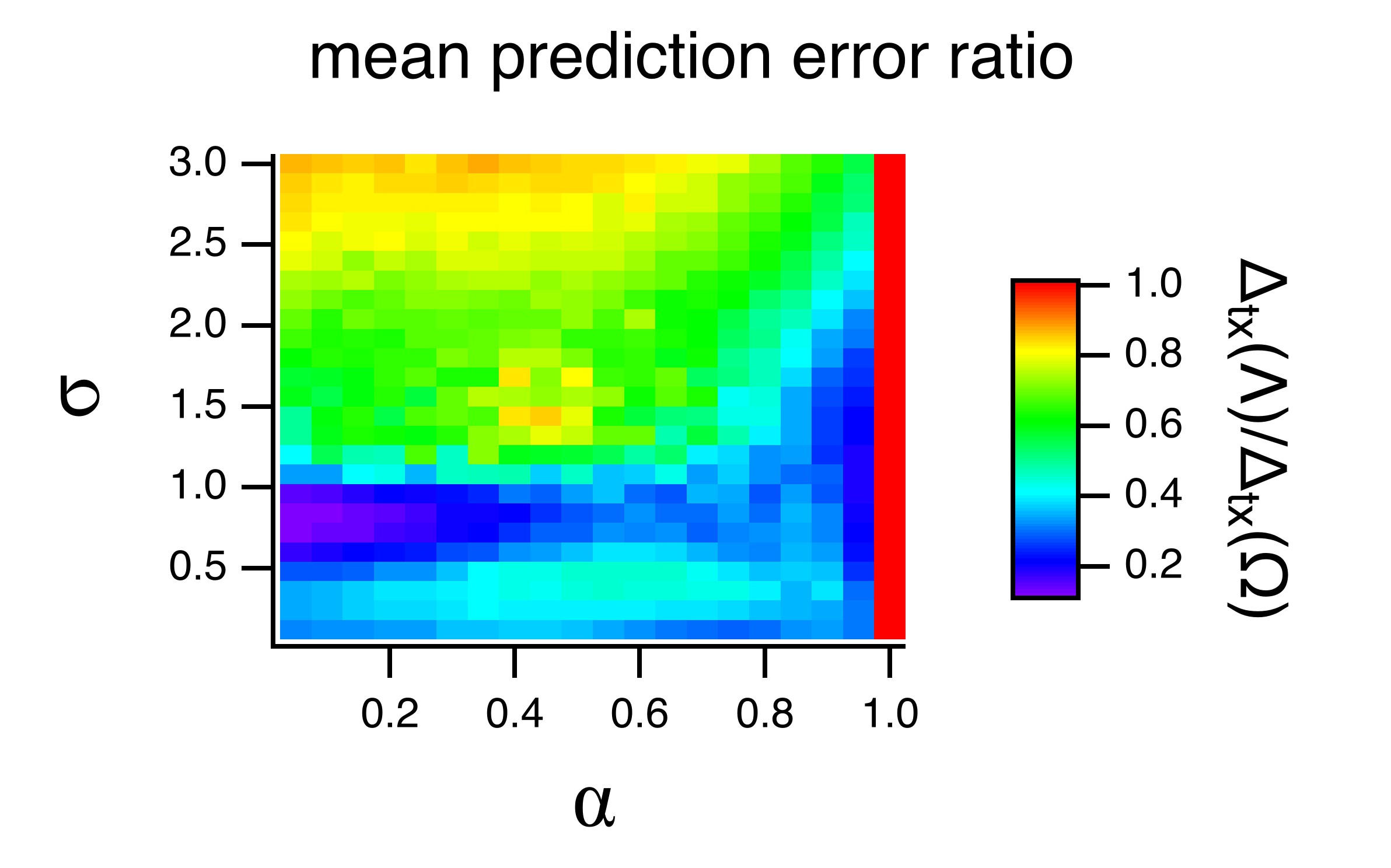} 
  \caption{ \label{umd_mean_xpred_ratio} Mean ratio of the prediction error for the Lorenz $x$ variable, $\Delta_{tx}(\Lambda)$, when five filters are used in the filter matrix $\Lambda$ to the prediction error for the leaky tanh reservoir by itself, as a function of the parameter $\alpha$ and the spectral radius $\sigma$. The reservoir had 80 nodes}
  \end{figure}

Figure \ref{umd_mean_xpred_ratio} shows the mean value of the ratio of the prediction error for the Lorenz $x$ signal with five filters following the reservoir to the prediction error for the reservoir only, as the parameter $\alpha$ and the spectral radius $\sigma$ are scanned. As with the error in fitting the Lorenz $z$ signal, the most improvement in predicting the $x$ signal when filters are added comes for the same parameter range for the smallest prediction error for the reservoir only.   
  
 \subsection{Laser System}
\begin{figure}[h]
\centering
\includegraphics[scale=0.8]{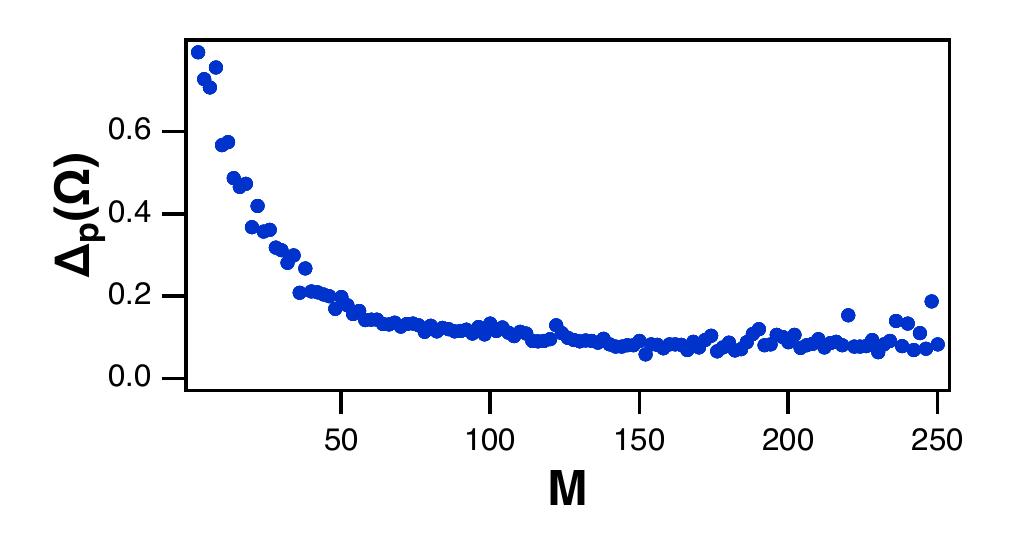} 
  \caption{ \label{laserpred} $\Delta_P(\Omega)$ is the error for predicting the future of the Lorenz $x$ signal  0.25 Lyapunov times into the future, plotted versus the number of nodes $M$. This prediction is for the reservoir computer only, using nodes modeled on the laser system (eqs. \ref{mzmap}-\ref{network}.}
  \end{figure}

 The prediction error for the laser system model is plotted in figure \ref{laserpred}. The prediction error saturates for somewhere between 50 and 100 nodes.
 
\begin{figure}[h]
\centering
\includegraphics[scale=0.8]{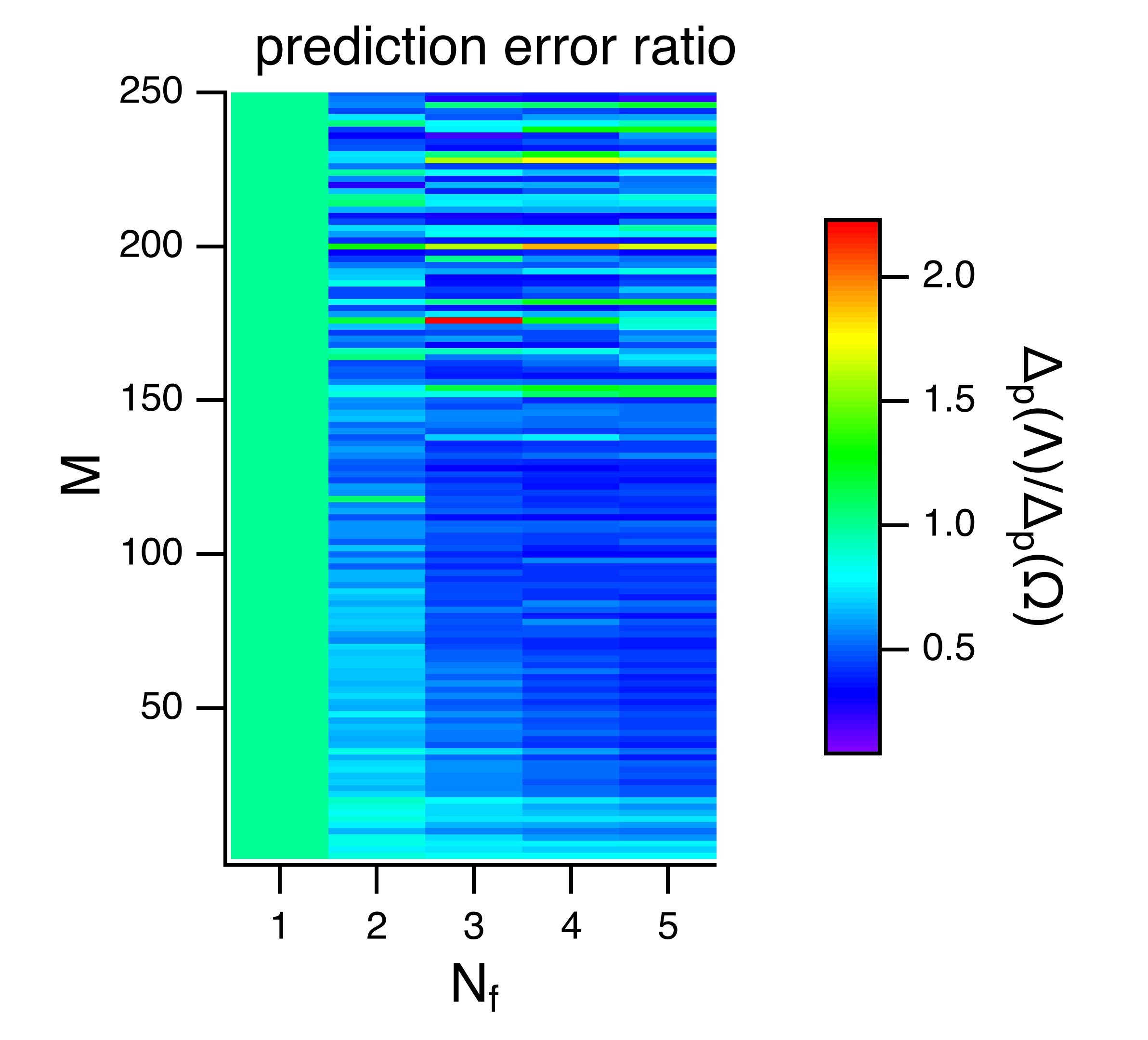} 
  \caption{ \label{laserpredfilt} Prediction error ratio $\Delta_P(\Lambda)/\Delta_P(\Omega)$ for predicting the future of the Lorenz $x$ signal when the reservoir computer based on the laser system model is augmented with up to five filters. The number of nodes is $M$, while $N_f$ is the number of filters.}
  \end{figure}

Adding filters to the reservoir computer modeled on the laser system can improve prediction, as shown in figure \ref{laserpredfilt}. The largest improvement in prediction error when three or more filters are used appears to come when the reservoir computer has more than 20 nodes. The prediction error does not improve for more than three filters. It is likely that the prediction error for the laser system is not limited by covariance rank in this reservoir computer, but rather by how well the reservoir signals $\chi_i(n)$ match the testing signal.

\section{Classification}
\label{class}
The reservoir computers from the previous sections will be used to determine if adding filters to a reservoir computer can improve the ability to classify a set of signals. The signals in this case are the $x$ component of the 19 Sprott chaotic systems \cite{sprott1994}. Each of the Sprott systems was numerically integrated with a time step of 0.5.

Some of the attractors for the Sprott systems have small basins of attraction, so rather than set random initial conditions to create different realizations of each Sprott system, a long time series of the $x$ signal was generated for each of the Sprott systems. The test and training signals were taken from different sections of this long time series. The reservoir computers used to classify the Sprott signals each had $M=100$ nodes.

The reservoir computers were trained with 100 training examples each. For each training example, the reservoir computer was first driven by a 1000 point signal to eliminate transients, after which the next 1000 output points from each node were used to fit the training signal. Fit coefficients were found using both the reservoir matrix $\Omega$ and the filter matrix $\Lambda$. For each of the Sprott systems, the fit coefficient vectors were given by ${\bf c}(j,k), j=1 \ldots 100, k=1 \ldots 19$, where $j$ indicated the $j$'th section of the $x$ signal and $k$ indicated the particular Sprott system. Each coefficient vector had $M$ components: ${\bf c}(j,k)=[c_1(j,k), c_2(j,k), \ldots c_M(j,k)]$, where $M$, the number of nodes, was 100. For each of the Sprott systems a reference coefficient vector was defined as the mean of the coefficient vectors:
\begin{equation}
\label{refcoeff}
{\bf{C}}\left( k \right) = \frac{1}{{100}}\sum\limits_{j = 1}^{100} {{\bf{c}}\left( {j,k} \right)}.
 \end{equation}
The set of ${\bf C}(k)=[C_1(k), C_2(k), \ldots C_M(k)], k=1 \ldots 19$ coefficients formed a reference library. 

To identify the Sprott systems, the reservoir computers were again driven with a 1000 point time series of the $x$ signal from each of the Sprott systems to eliminate transients. The next 1000 points were saved in the reservoir computer matrix $\Omega$ or the filter matrix $\Lambda$.  Once again, for each section a set of fit coefficients ${\bf c}(j,l), j=1 \ldots 1000, l=1 \ldots 19$ was found, for both the reservoir computer matrix and the filter matrix. 

Each time a coefficient vector was found, it was compared to the reference library according to
\begin{equation}
\label{refdiff}
{\Psi _j}\left( {l,k} \right) = \sqrt {\sum\limits_{i = 1}^M {{{\left[ {{c_i}\left( {j,l} \right) - {C_j}\left( k \right)} \right]}^2}} } 
\end{equation}
where $i=1 \ldots M$ indicated the components (or node numbers) of the coefficient vectors. The difference $\Psi_j(l,k)$ was computed for all 19 reference coefficient vectors ${\bf C}_k$, and the value of $k$ that gave the minimum of $\Psi_j(l,k)$ was identified as the Sprott system that generated the coefficient vector ${\bf c}(i,l)$.

\begin{figure}[h]
\centering
\includegraphics[scale=0.8]{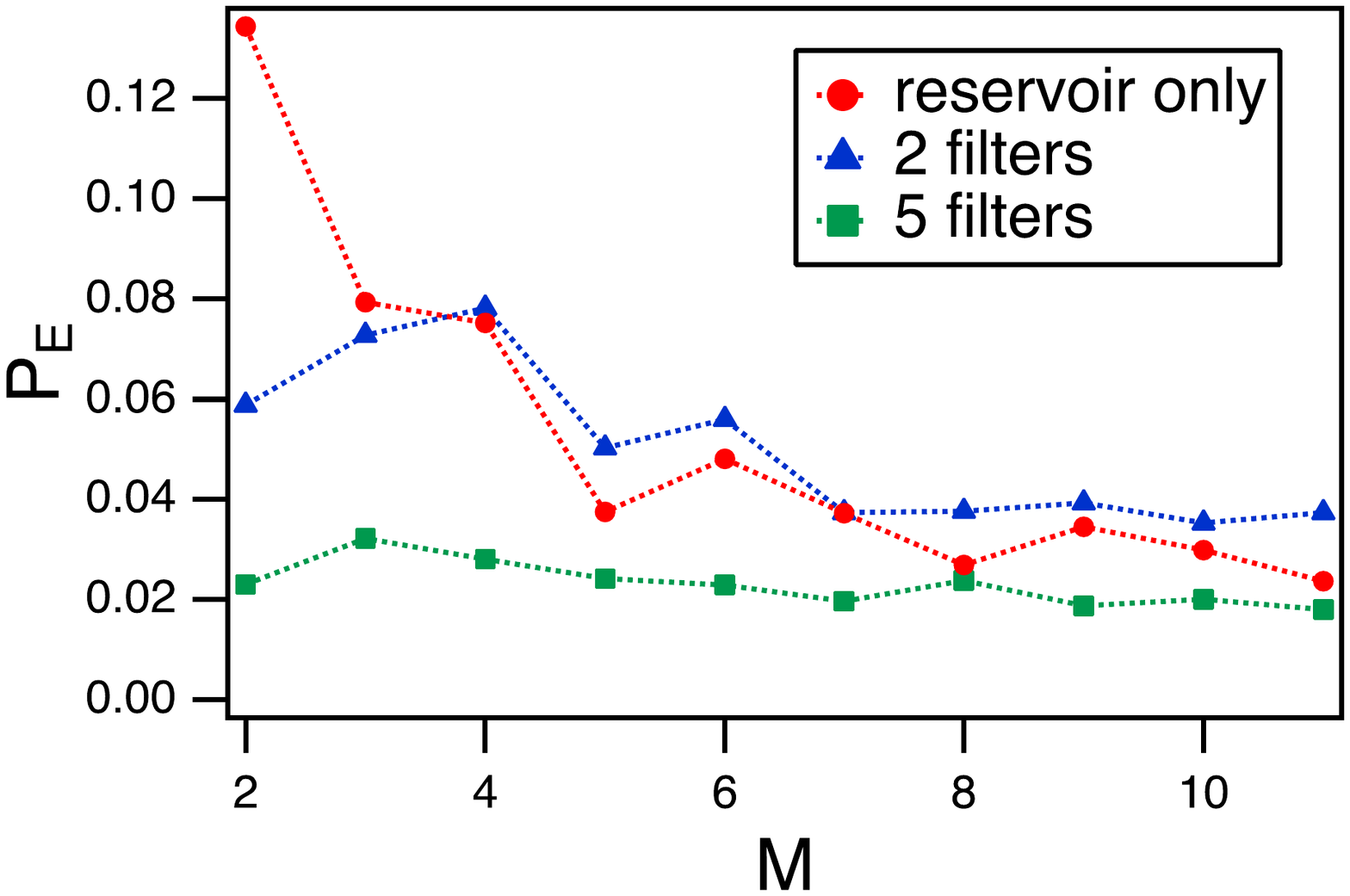} 
  \caption{ \label{umdclass} Probability of error $P_E$ in identifying the source of an $x$ signal from one of the 19 Sprott systems as a function of the number of nodes $M$ in the leaky tanh reservoir computer.}
  \end{figure}

The probability of making an error when identifying from which of the Sprott systems an $x$ signal originated is shown in figure \ref{umdclass}. Each time a signal from a Sprott system was compared to the reference library, if the value of $k$ that gave a minimum of $\Psi_j(l,k)$ did not correspond to the Sprott system that generated the signal, an error was recorded. The probability of error $P_E$ was the total number of errors divided by the total number of comparisons.

  Figure \ref{umdclass} shows that adding just two filters (really just one filter, since one of the filters is the identity) to the leaky tanh reservoir computer improves the ability to identify the Sprott systems if the reservoir computer has only two nodes, but not if it has more than two nodes. Adding five filters improves the classification of signals if the reservoir computer has less than eight nodes. Once the reservoir has eight nodes, adding more nodes or adding filters gives little extra benefit.

  Adding filters to a reservoir computer showed a greater advantage when fitting signals (Section \ref{fitting}) than when classifying signals. In fitting signals the reservoir time series acts as a basis, so the covariance rank of the reservoir output is important. Adding filters increases this rank. Classifying signals may not depend as much on how well the reservoir computer fits the signals; what is more important is that the set of fit coefficients are sufficiently different for different inputs. Adding filters to a reservoir computer does create more coefficients, but the filters are linear, so the extra coefficients may not be useful in distinguishing the different Sprott signals.

\begin{figure}[h]
\centering
\includegraphics[scale=0.8]{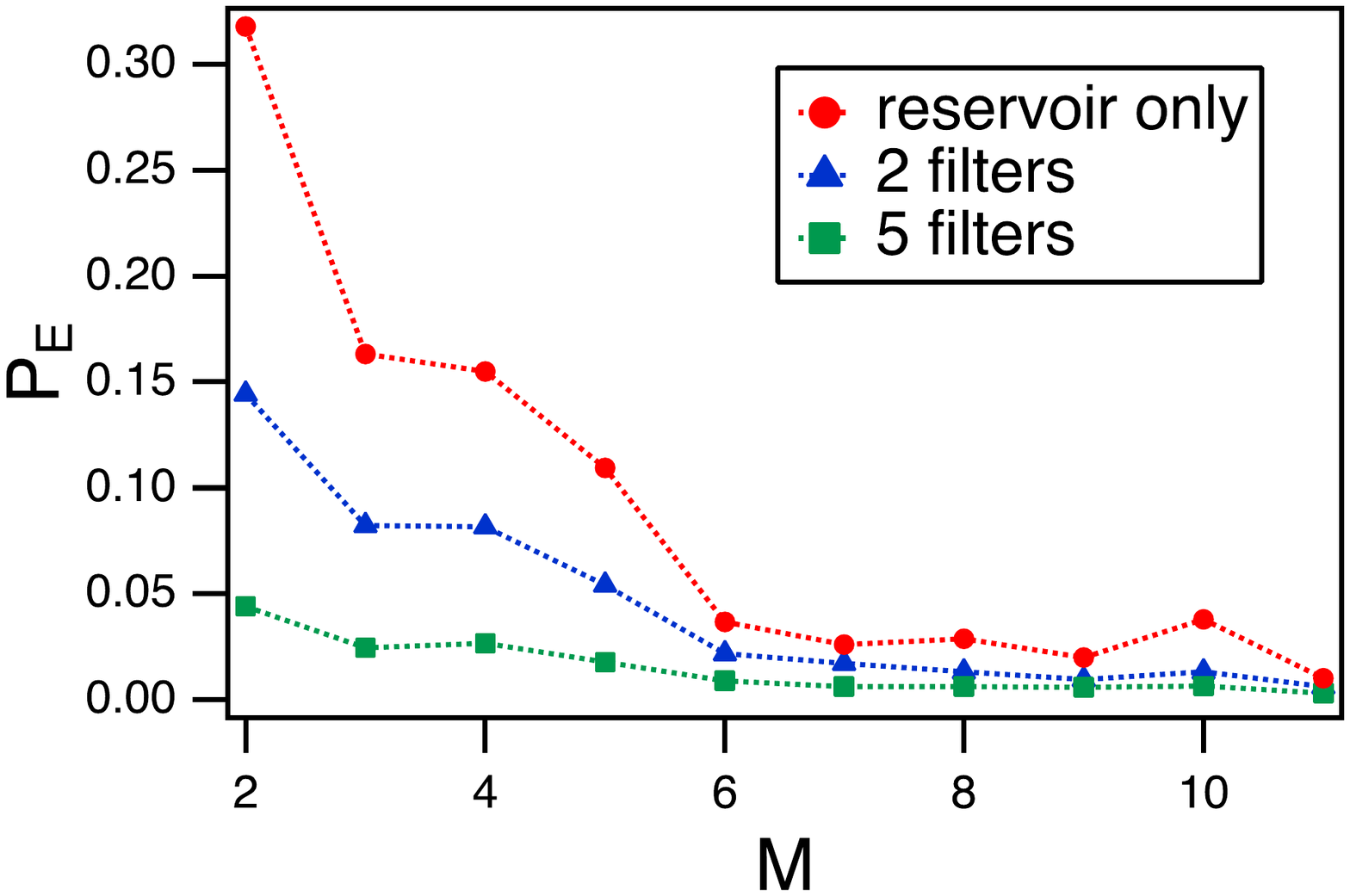} 
  \caption{ \label{laserclass} Probability of error $P_E$ in identifying the source of an $x$ signal from one of the 19 Sprott systems as a function of the number of nodes $M$ in the reservoir computer based on the laser system model of eqs. (\ref{mzmap}-\ref{network}).}
  \end{figure}

Figure \ref{laserclass} shows the probability of error in identifying the Sprott systems using a reservoir computer based on the laser system model of eqs. (\ref{mzmap}-\ref{network}).  

  Figure \ref{laserclass} for the laser model reservoir computer shows that adding two or five filters to this type of reservoir computer does reduce classification error when the reservoir has less than six nodes. Once the reservoir computer has six or more nodes, adding filters does not show any advantage for classification.
 
     \begin{figure}[h]
\centering
\includegraphics[scale=0.8]{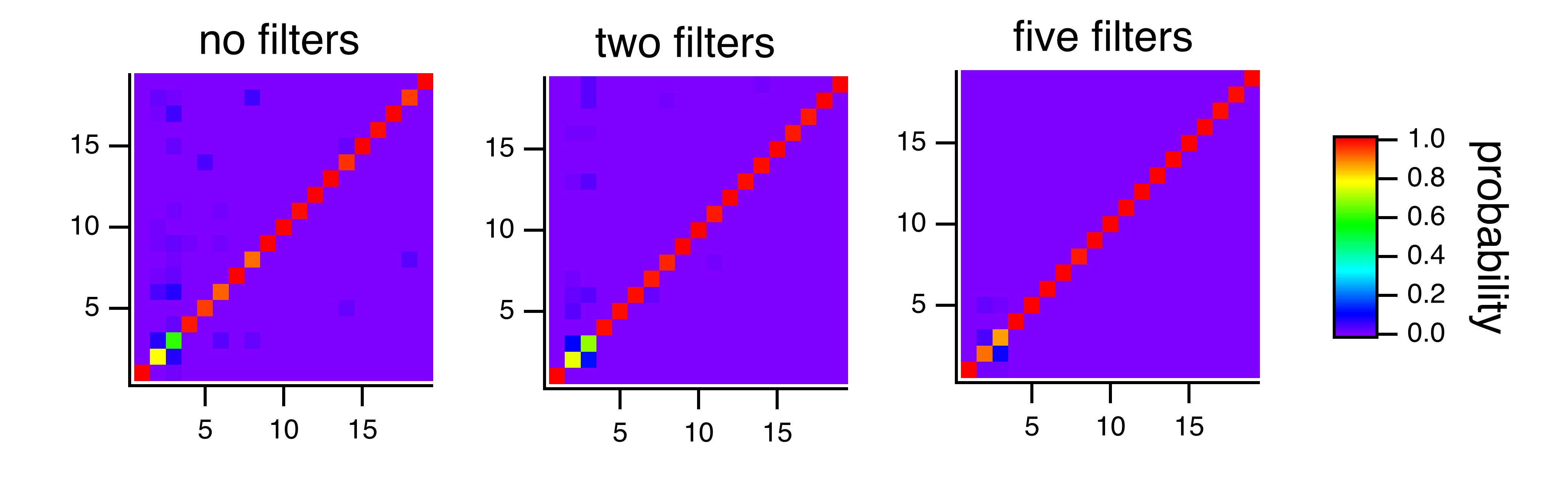} 
  \caption{ \label{umdconfmat} Confusion matrices for classifying the 19 Sprott chaotic systems based on the leaky tanh reservoir computer. The reservoir computer had four nodes for these figures. }
  \end{figure}   
   
        \begin{figure}[h]
\centering
\includegraphics[scale=0.8]{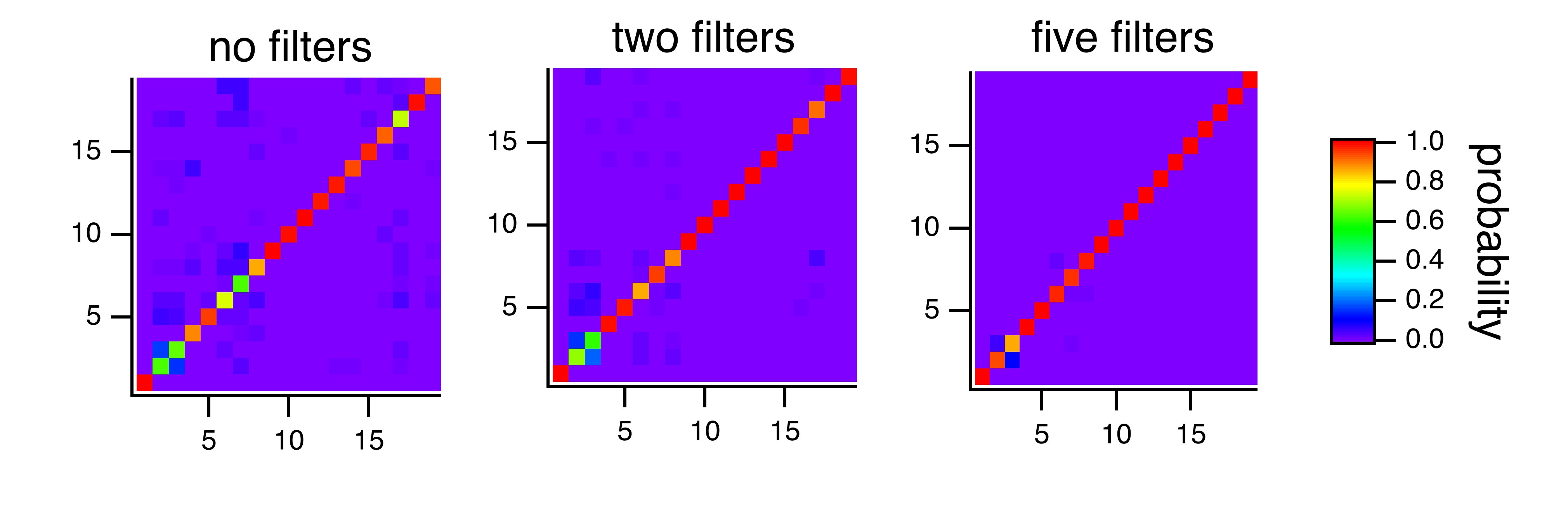} 
  \caption{ \label{mzconfmat} Confusion matrices for classifying the 19 Sprott chaotic systems based on the laser reservoir computer. The reservoir computer had four nodes for these figures. }
  \end{figure}  
   
 Adding filters to a very small reservoir computer can increase the dimension of the coefficient vector, but once the coefficient vector has enough dimensions, adding additional filters does not lower the classification error.  Still, if creating a reservoir with many coupled nodes is difficult or expensive, adding linear filters to a small reservoir computer can be useful.
  
Figures \ref{umdconfmat} and \ref{mzconfmat} are confusion matrices for the Sprott classification problem for the leaky tanh reservoir computer or the laser reservoir computer. In each of these figures, the reservoir computer had four nodes, so when two filters were added there were 8 coefficients (eq. \ref{refcoeff}) and when five filters were added there were 20 coefficients.

Figure \ref{umdconfmat} shows that for the leaky tanh reservoir computer, the classification accuracy was limited by systems 2 and 3. The reservoir computer with two added filters has fewer misclassifications than for the reservoir computer alone, but the probability of misclassification between systems 2 and 3 is about the same. When five filters are added, the probability of misclassification between systems 2 and 3 is smaller.

With no added filters, the laser system reservoir computer confusion matrices in figure \ref{mzconfmat} also show a high probability of misclassification between systems 2 and 3, but also a high probability of misclassification for systems 6, 7, 8 and 17. The classification errors for all but systems 2 and 3 drop sharply when two filters are added, and the misclassification between systems 2 and 3 is smaller when five filters are added.

\section{Memory}
\label{mem_cal}
Memory capacity, as defined in \cite{jaeger2002}, is considered to be an important quantity in reservoir computers. Memory capacity is a measure of how well the reservoir can reproduce previous values of the input signal.

The memory capacity as a function of delay is
\begin{equation}
\label{memdel}
{\rm{M}}{{\rm{C}}_k} = \frac{{\sum\limits_{n = 1}^N {\left[ {s\left( {n - k} \right) - \overline s } \right]\left[ {{g_k}\left( n \right) - \overline {{g_k}} } \right]} }}{{\sum\limits_{n = 1}^N {\left[ {s\left( {n - k} \right) - \overline s } \right]\sum\limits_{n = 1}^N {\left[ {{g_k}\left( n \right) - \overline {{g_k}} } \right]} } }}
\end{equation}
where the overbar indicator indicates the mean. The signal $g_k(n)$ is the fit of the reservoir signals $\chi_i(n)$ to the delayed input signal $s(n-k)$. The memory capacity is
\begin{equation}
\label{memcap}
{\rm{MC}} = \sum\limits_{k = 1}^\infty  {{\rm{M}}{{\rm{C}}_k}} 
\end{equation}

Input signals such as the Lorenz $x$ signal contain correlations in time, which will cause errors in the memory calculation, so in eq. (\ref{memdel}), $s(n)$ is a random signal uniformly distributed between -1 and +1. There are some drawbacks to defining memory in this way; the reservoir is nonlinear, so its response will be different for different input signals, but this memory definition is the standard definition used in the field of reservoir computing.

\begin{figure}[h]
\centering
\includegraphics[scale=0.8]{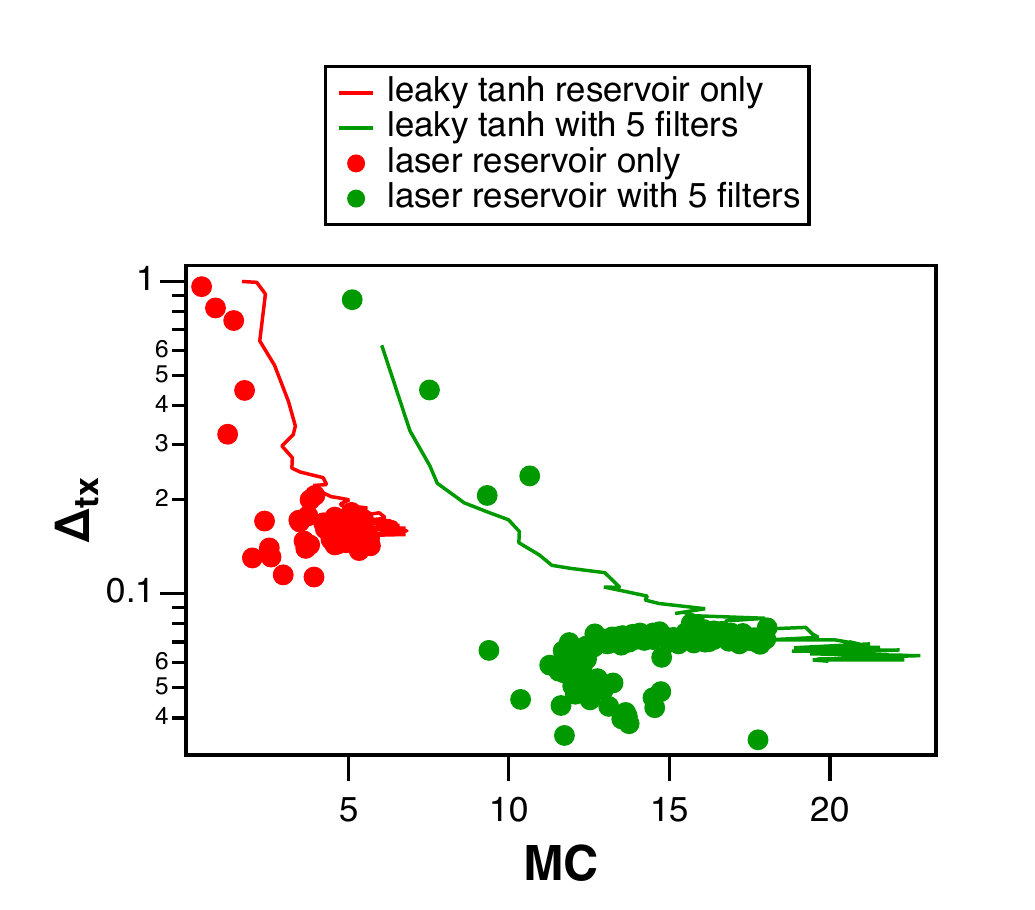} 
  \caption{ \label{caprank} Testing error $\Delta_{tx}$ for both types of reservoir as a function of memory capacity}
  \end{figure}

Figure \ref{caprank} shows the testing error for both reservoir types as a function of memory capacity.

Figure \ref{caprank} shows two things; the testing error decreases as memory capacity increases, and the memory capacity for both reservoirs is higher with five filters following the reservoir than for the reservoir only. The data is more scattered for the laser system, but the trend is still there. It has been noted that there is a tradeoff between nonlinearity and memory in reservoir computers \cite{inubushi2017}; adding filters is a way to add memory to a reservoir computer without affecting the nonlinearity.

\section{Summary}
Reservoir computers should show the greatest advantage over other types of computing when they are built as analog systems, but building these systems may be difficult or expensive. Creating individual nonlinear nodes may be difficult, but the largest cost in building analog reservoir computers may be in connecting the nodes in a network. The work in this paper demonstrates that reservoir computers may be expanded if the nonlinear network is followed by a set of linear filters. The design and implementation of linear filters is well known, so there should be little cost for adding filters to the reservoir computer.

This work showed that adding filters improved the performance of two types of reservoir computer for signal fitting, for prediction and for classification. Improvements in classifying signals were largest for small reservoir computers, but small reservoir computers are where the improvement is most needed.

Reservoir computers may be expanded using other types of functions besides linear FIR filters; the linear filters were used here because they are simple to design and characterize. It is even possible to use nonlinear functions; one early paper, for example, connected linear nodes into a network and followed the linear network with nonlinear output functions \cite{boyd1985}.
  
  This work was supported by the Naval Research Laboratory's Basic Research Program.


\begin{thebibliography}{10}
\expandafter\ifx\csname url\endcsname\relax
  \def\url#1{\texttt{#1}}\fi
\expandafter\ifx\csname urlprefix\endcsname\relax\def\urlprefix{URL }\fi
\expandafter\ifx\csname href\endcsname\relax
  \def\href#1#2{#2} \def\path#1{#1}\fi

\bibitem{jaeger2001}
H.~Jaeger, The echo state approach to analysing and training recurrent neural
  networks-with an erratum note, German National Research Center for
  Information Technology GMD Technical Report 148~(1) (2001) 34.
\newblock \href
  {http://dx.doi.org/http://publica.fraunhofer.de/documents/B-73135.html}
  {\path{doi:http://publica.fraunhofer.de/documents/B-73135.html}}.

\bibitem{natschlaeger2002}
T.~Natschlaeger, W.~Maass, H.~Markram, The "liquid computer": A novel strategy
  for real-time computing on time series, Special Issue on Foundations of
  Information Processing of TELEMATIK 8~(1) (2002) 39--43.

\bibitem{lu2017}
Z.~Lu, J.~Pathak, B.~Hunt, M.~Girvan, R.~Brockett, E.~Ott, Reservoir observers:
  Model-free inference of unmeasured variables in chaotic systems, Chaos: An
  Interdisciplinary Journal of Nonlinear Science 27~(4) (2017) 041102.
\newblock \href {http://dx.doi.org/10.1063/1.4979665}
  {\path{doi:10.1063/1.4979665}}.

\bibitem{larger2012}
L.~Larger, M.~C. Soriano, D.~Brunner, L.~Appeltant, J.~M. Gutierrez,
  L.~Pesquera, C.~R. Mirasso, I.~Fischer, Photonic information processing
  beyond turing: an optoelectronic implementation of reservoir computing,
  Optics Express 20~(3) (2012) 3241--3249.
\newblock \href {http://dx.doi.org/10.1364/oe.20.003241}
  {\path{doi:10.1364/oe.20.003241}}.

\bibitem{van_der_sande2017}
G.~V. der Sande, D.~Brunner, M.~C. Soriano, Advances in photonic reservoir
  computing, Nanophotonics 6~(3) (2017) 561--576.
\newblock \href {http://dx.doi.org/10.1515/nanoph-2016-0132}
  {\path{doi:10.1515/nanoph-2016-0132}}.

\bibitem{schurmann2004}
F.~Schurmann, K.~Meier, J.~Schemmel, Edge of chaos computation in mixed-mode
  vlsi - a hard liquid, in: Advances in Neural Information Processing Systems
  17, MIT Press, 2004, pp. 1201--1208.

\bibitem{dion2018}
G.~Dion, S.~Mejaouri, J.~Sylvestre, Reservoir computing with a single
  delay-coupled non-linear mechanical oscillator, Journal of Applied Physics
  124~(15) (2018) 152132.
\newblock \href {http://dx.doi.org/10.1063/1.5038038}
  {\path{doi:10.1063/1.5038038}}.

\bibitem{canaday2018}
D.~Canaday, A.~Griffith, D.~J. Gauthier, Rapid time series prediction with a
  hardware-based reservoir computer, Chaos: An Interdisciplinary Journal of
  Nonlinear Science 28~(12) (2018) 123119.
\newblock \href {http://dx.doi.org/10.1063/1.5048199}
  {\path{doi:10.1063/1.5048199}}.

\bibitem{carroll2019}
T.~L. Carroll, L.~M. Pecora, Network structure effects in reservoir computers,
  Chaos 29~(8) (2019) 083130.
\newblock \href {http://dx.doi.org/10.1063/1.5097686}
  {\path{doi:10.1063/1.5097686}}.

\bibitem{carroll2020}
T.~L. Carroll, Dimension of reservoir computers, Chaos: An Interdisciplinary
  Journal of Nonlinear Science 30~(1) (2020) 013102.
\newblock \href {http://dx.doi.org/10.1063/1.5128898}
  {\path{doi:10.1063/1.5128898}}.

\bibitem{dambre2012}
J.~Dambre, D.~Verstraeten, B.~Schrauwen, S.~Massar, Information processing
  capacity of dynamical systems, Scientific Reports 2 (2012) 514.
\newblock \href {http://dx.doi.org/10.1038/srep00514
  https://www.nature.com/articles/srep00514#supplementary-information}
  {\path{doi:10.1038/srep00514
  https://www.nature.com/articles/srep00514#supplementary-information}}.

\bibitem{badii1988}
R.~Badii, G.~Broggi, B.~Derighetti, M.~Ravani, S.~Ciliberto, A.~Politi, M.~A.
  Rubio, Dimension increase in filtered chaotic signals, Phys Rev Lett 60~(11)
  (1988) 979--982.
\newblock \href {http://dx.doi.org/10.1103/PhysRevLett.60.979}
  {\path{doi:10.1103/PhysRevLett.60.979}}.

\bibitem{fpga2020}
Wikipedia,
  \href{https://en.wikipedia.org/wiki/Field-programmable_gate_array}{Field-programmable
  gate array}\href
  {http://dx.doi.org/https://en.wikipedia.org/wiki/Field-programmable_gate_array}
  {\path{doi:https://en.wikipedia.org/wiki/Field-programmable_gate_array}}.
\newline\urlprefix\url{https://en.wikipedia.org/wiki/Field-programmable_gate_array}

\bibitem{tietze1991}
U.~Tietze, C.~Shenk, Electronic Circuits, Springer, Berlin, 1991.

\bibitem{jaeger2007}
H.~Jaeger, M.~Luko{\v s}evi{\v c}ius, D.~Popovici, U.~Siewert, Optimization and
  applications of echo state networks with leaky- integrator neurons, Neural
  Networks 20~(3) (2007) 335--352.
\newblock \href
  {http://dx.doi.org/https://doi.org/10.1016/j.neunet.2007.04.016}
  {\path{doi:https://doi.org/10.1016/j.neunet.2007.04.016}}.

\bibitem{joliffe2011}
I.~T. Jolliffe, Principal component analysis, Springer, 2011.

\bibitem{lorenz1963}
E.~N. Lorenz, Deterministic non-periodic flow, Journal of Atmospheric Science
  20~(2) (1963) 130--141.
\newblock \href
  {http://dx.doi.org/10.1175/1520-0469(1963)020<0130:DNF>2.0.CO;2}
  {\path{doi:10.1175/1520-0469(1963)020<0130:DNF>2.0.CO;2}}.

\bibitem{jaeger2004}
H.~Jaeger, H.~Haas, Harnessing nonlinearity: Predicting chaotic systems and
  saving energy in wireless communication, Science 304~(5667) (2004) 78--80.
\newblock \href {http://dx.doi.org/10.1126/science.1091277}
  {\path{doi:10.1126/science.1091277}}.

\bibitem{sprott1994}
J.~C. Sprott, Some simple chaotic flows, Physical Review E 50~(2) (1994)
  R647--R650.
\newblock \href {http://dx.doi.org/10.1103/PhysRevE.50.R647}
  {\path{doi:10.1103/PhysRevE.50.R647}}.

\bibitem{jaeger2002}
H.~Jaeger, Short term memory in echo state networks, Technical report
  GMD-Forschungszentrum Informationstechnik.

\bibitem{inubushi2017}
M.~Inubushi, K.~Yoshimura, Reservoir computing beyond memory-nonlinearity
  trade-off, Scientific Reports 7~(1) (2017) 10199.
\newblock \href {http://dx.doi.org/10.1038/s41598-017-10257-6}
  {\path{doi:10.1038/s41598-017-10257-6}}.

\bibitem{boyd1985}
S.~Boyd, L.~Chua, Fading memory and the problem of approximating nonlinear
  operators with volterra series, IEEE Transactions on Circuits and Systems
  32~(11) (1985) 1150--1161.
\newblock \href {http://dx.doi.org/10.1109/TCS.1985.1085649}
  {\path{doi:10.1109/TCS.1985.1085649}}.

\end{thebibliography}
\end{document}